\documentclass{article}

     \usepackage[preprint]{neurips_2020}

\usepackage[utf8]{inputenc} 
\usepackage[T1]{fontenc}    
\usepackage{hyperref}       
\usepackage{url}            
\usepackage{booktabs}       
\usepackage{amsfonts}       
\usepackage{nicefrac}       
\usepackage{microtype}      
\usepackage{graphicx}
\usepackage{amsmath}
\usepackage{float}
\usepackage{bbm}
\usepackage{bm}

\title{Fast differentiable DNA and protein sequence optimization for molecular design}

\author{%
  Johannes Linder\thanks{All code available at http://www.github.com/johli/seqprop}\\
  Paul G. Allen School\\
  of Computer Science \& Engineering\\
  University of Washington\\
  Seattle, WA 98195, USA \\
  \texttt{jlinder2@cs.washington.edu} \\
  \And
  Georg Seelig\thanks{Affiliated with Department of Electrical \& Computer Engineering, University of Washington}\\
  Paul G. Allen School\\
  of Computer Science \& Engineering\\
  University of Washington\\
  Seattle, WA 98195, USA \\
}

\begin{document}

\maketitle

\begin{abstract}
  Designing DNA and protein sequences with improved function has the potential to greatly accelerate synthetic biology. Machine learning models that accurately predict biological fitness from sequence are becoming a powerful tool for molecular design. Activation maximization offers a simple design strategy for differentiable models: one-hot coded sequences are first approximated by a continuous representation which is then iteratively optimized with respect to the predictor oracle by gradient ascent. While elegant, this method suffers from vanishing gradients and may cause predictor pathologies leading to poor convergence. Here, we build on a previously proposed straight-through approximation method to optimize through discrete sequence samples. By normalizing nucleotide logits across positions and introducing an adaptive entropy variable, we remove bottlenecks arising from overly large or skewed sampling parameters. The resulting algorithm, which we call Fast SeqProp, achieves up to 100-fold faster convergence compared to previous versions of activation maximization and finds improved fitness optima for many applications. We demonstrate Fast SeqProp by designing DNA and protein sequences for six deep learning predictors, including a protein structure predictor.
\end{abstract}

\section{Introduction}

Rational design of DNA and protein sequences enables rapid development of novel drug molecules, vaccines, biological circuits and more. Most design methods are guided by predictive models that reliably relate sequence to fitness or function. In recent years, these models are often based on deep learning (Eraslan et al., 2019; Zou et al., 2019; Tareen et al., 2019). For example, neural networks have been trained to predict transcription factor binding (Alipanahi et al., 2015; Avsec et al., 2019), chromatin modifications (Zhou et al., 2015) and transcriptional activity (Movva et al., 2019). At the level of RNA, they have been used to predict translation (Sample et al., 2019; Karollus et al., 2020), splicing (Jaganathan et al., 2019; Cheng et al., 2019) and polyadenylation (Bogard et al., 2019; Arefeen et al., 2019; Li et al., 2020). Neural networks have even been used to predict protein structure (AlQuraishi 2019, Senior et al., 2020; Yang et al., 2020).

A wide range of methods have previously been applied to computational sequence design, for example Genetic Algorithms (Deaton et al., 1996), Simulated Annealing (Hao et al., 2015), Bayesian optimization (Belanger et al., 2019) and population-based methods (Xiao et al. 2009; Ibrahim et al., 2011; Mustaza et al., 2011; Angermueller et al., 2020). More recently, methods combining adaptive sampling or other optimization techniques with deep generative networks have been used to model distributions of sequences with desired functional properties (Gómez-Bombarelli et al., 2018; Gupta et al., 2019; Brookes et al., 2019; Yang et al., 2019; Riesselman et al., 2019; Costello et al., 2019, Linder et al. 2020), including Deep Exploration Networks (DENs), developed in earlier work by our group. While powerful, these methods first require selecting an appropriate generative network and tuning several hyper-parameters. 

Perhaps the simplest approach to sequence design based on a differentiable fitness predictor is to optimize the input pattern by gradient ascent (Lanchantin et al., 2016; Killoran et al., 2017; Gómez-Bombarelli et al., 2018; Bogard et al., 2019). The method, commonly referred to as activation maximization, uses the gradient of the neural network output to make incremental changes to the input. While simple, activation maximization cannot be directly applied to sequences as they are discrete and not amenable to gradient ascent. Several extensions have been proposed to rectify this; Killoran et al., (2017) used a softmax layer to turn the sequences into continuous relaxations, and in previous work we developed \textit{SeqProp} which used straight-through gradients to optimize discrete samples (Bogard et al., 2019). However, as our results indicate below, both methods converge slowly. Furthermore, continuous input relaxations may cause predictor pathologies leading to poor designs.

Here, we combine discrete nucleotide sampling and straight-through approximation with normalization across the parameters -- nucleotide logits -- of the sampling distributions, resulting in a markedly improved sequence design method which we refer to as \textit{Fast SeqProp}. We apply Fast SeqProp to a range of DNA and protein design tasks including the design of strong enhancers, 5'UTRs, alternative polyadenylation signals and protein structures. We demonstrate up to a 100-fold optimization speedup, and improved optima, for these design tasks compared to prior methods based on activation maximization. We independently validate designed sequences by scoring them with models not used in the optimization procedure. We further show that our method can outperform global search heuristics such as Simulated Annealing and more recent sampling-based methods based on generative models. Unlike the latter approaches, Fast SeqProp does not require training of an independent generator. It is thus model- and parameter-free, making it easy to use when designing smaller sequence sets. Moreover, Fast SeqProp can incorporate many different regularization techniques to design sequences that are not too distant from the original training data and thus maintain confidence, such as regularization based on a variational autoencoder (VAE) and optimization of probabilistic predictor models which are capable of estimating uncertainty.

\section{Background}

Given a sequence-predictive neural network $\mathcal{P}$ and an initial input pattern $\bm{x}^{(0)}$, the gradient ascent method seeks to maximize the predicted \textit{fitness} $\mathcal{P}(\bm{x}) \in \mathbb{R}$ by tuning the input pattern $\bm{x}$:
\begin{align} \label{eq:1}
  \max_{\bm{x}} \enskip \mathcal{P}(\bm{x})
\end{align}
Assuming $\mathcal{P}$ is differentiable, we can compute the gradient $\nabla_{\bm{x}} \mathcal{P}(\bm{x})$ with respect to the input and optimize $\bm{x}$ by updating the variable in the direction of the fitness gradient (Simonyan et al., 2013):
\begin{align} \label{eq:2}
    \bm{x}^{(t+1)} \leftarrow \bm{x}^{(t)} + \eta \cdot \nabla_{\bm{x}} \mathcal{P}(\bm{x})
\end{align}

\noindent However, sequences are usually represented as one-hot coded patterns ($\bm{x} \in \{0, 1\}^{N \times M}$, where $N$ is the sequence length and $M$ the number of channels; $M=4$ for nucleic acids and $M=20$ for proteins), and discrete variables cannot be optimized by gradient ascent. Several different reparameterizations of $\bm{x}$ have been proposed to bypass this issue. In one of the earliest implementations, Lanchantin et al. (2016) represented the sequence as an unstructured, real-valued pattern ($\bm{x} \in \mathbb{R}^{N \times M}$) but imposed an L$2$-penalty on $\bm{x}$ in order to keep it from growing too large and causing predictor pathologies. After optimization, this real-valued pattern is interpreted as a sequence logo from which samples can be drawn. However, the method was introduced mainly as a visualization tool rather than a sequence design approach. 
Killoran et al. (2017) later introduced a softmax reparameterization, turning $\bm{x}$ into a continuous relaxation $\sigma(\bm{l})$:
\begin{align} \label{eq:3}
    \sigma(\bm{l})_{ij} = \frac{ e^{\bm{l}_{ij}} }{ \sum_{k=1}^{4} e^{\bm{l}_{ik}} }
\end{align}
Here $\bm{l}_{ij} \in \mathbb{R}$ are differentiable \textit{nucleotide logits}. The gradient of $\sigma(\bm{l})$ with respect to $\bm{l}$ is defined as:
\begin{align} \label{eq:4}
    \frac{\partial \sigma(\bm{l})_{ij}}{\partial \bm{l}_{ik}} = \sigma(\bm{l})_{ik} \cdot \left(\mathbbm{1}_{(j = k)} - \sigma(\bm{l})_{ij}\right)
\end{align}
Given Equation \ref{eq:3} and \ref{eq:4}, we can maximize $\mathcal{P}(\sigma(\bm{l}))$ with respect to the logits $\bm{l}$ using the gradient $\nabla_{\bm{l}} \mathcal{P}\left(\sigma(\bm{l})\right)$. While elegant, there are two issues with this architecture. First, the gradient in Equation \ref{eq:4} becomes vanishingly small for large values of $\bm{l}_{ij}$ (when $\sigma(\bm{l})_{ik} \approx 0$ or $\sigma(\bm{l})_{ij} \approx 1$), halting convergence. Second, sequence-predictive neural networks have only been trained on discrete one-hot coded patterns and the predictive power of $\mathcal{P}$ may be poor on a continuous relaxation such as $\sigma(\bm{l})$.

Following advances in gradient estimators for discretized neurons (Bengio, Léonard and Courville, 2013; Courbariaux et al., 2016), we developed \textit{SeqProp}, a version of activation maximization that replaces the softmax transform $\sigma$ with a discrete, stochastic sampler $\delta$ (Bogard et al., 2019):
\begin{align} \label{eq:5}
    \delta(\bm{l})_{ij} = \mathbbm{1}_{(Z_{i} = j)}
\end{align}
Here, $Z_{i} \sim \sigma(\bm{l})_{i}$ is a randomly drawn categorical nucleotide at the $i$:th position from the (softmax) probability distribution $\sigma(\bm{l})_{i}$. The nucleotide logits $\bm{l}_{ij}$ are thus used as parameters to $N$ categorical distributions, from which we sample nucleotides $\{Z_{i}\}_{i=1}^{N}$ and construct a discrete, one-hot coded pattern $\delta(\bm{l}) \in \{0, 1\}^{N \times M}$. While $\delta(\bm{l})$ is not directly differentiable, $\bm{l}$ can be updated based on the estimate of $\nabla_{\bm{l}} \mathcal{P}(\delta(\bm{l}))$ using straight-through (ST) approximation. Rather than using the original ST estimator of (Bengio et al. 2013), we here adopt an estimator with theoretically better properties from (Chung et al., 2016) where the gradient of $\delta(\bm{l})_{ij}$ is replaced by that of the softmax $\sigma(\bm{l})_{ij}$:
\begin{align} \label{eq:6}
    \frac{\partial \delta(\bm{l})_{ij}}{\partial \bm{l}_{ik}} = \frac{\partial \sigma(\bm{l})_{ij}}{\partial \bm{l}_{ik}} = \sigma(\bm{l})_{ik} \cdot (\mathbbm{1}_{(j = k)} - \sigma(\bm{l})_{ij})
\end{align}

Using discrete samples as input to $\mathcal{P}$ removes any pathologies from continuous input relaxations. But, as we show below, convergence remains almost as slow as the softmax method. Switching to the original ST estimator ($\frac{\partial \delta(\bm{l})_{ij}}{\partial \bm{l}_{ij}} = 1$) speeds up convergence, but results in worse fitness optima (see Supplementary Information, Figure S1G for a comparison).

\section{Fast Stochastic Sequence Backpropagation}

\begin{figure}[H]
  \centering
  \includegraphics[scale=1.0]{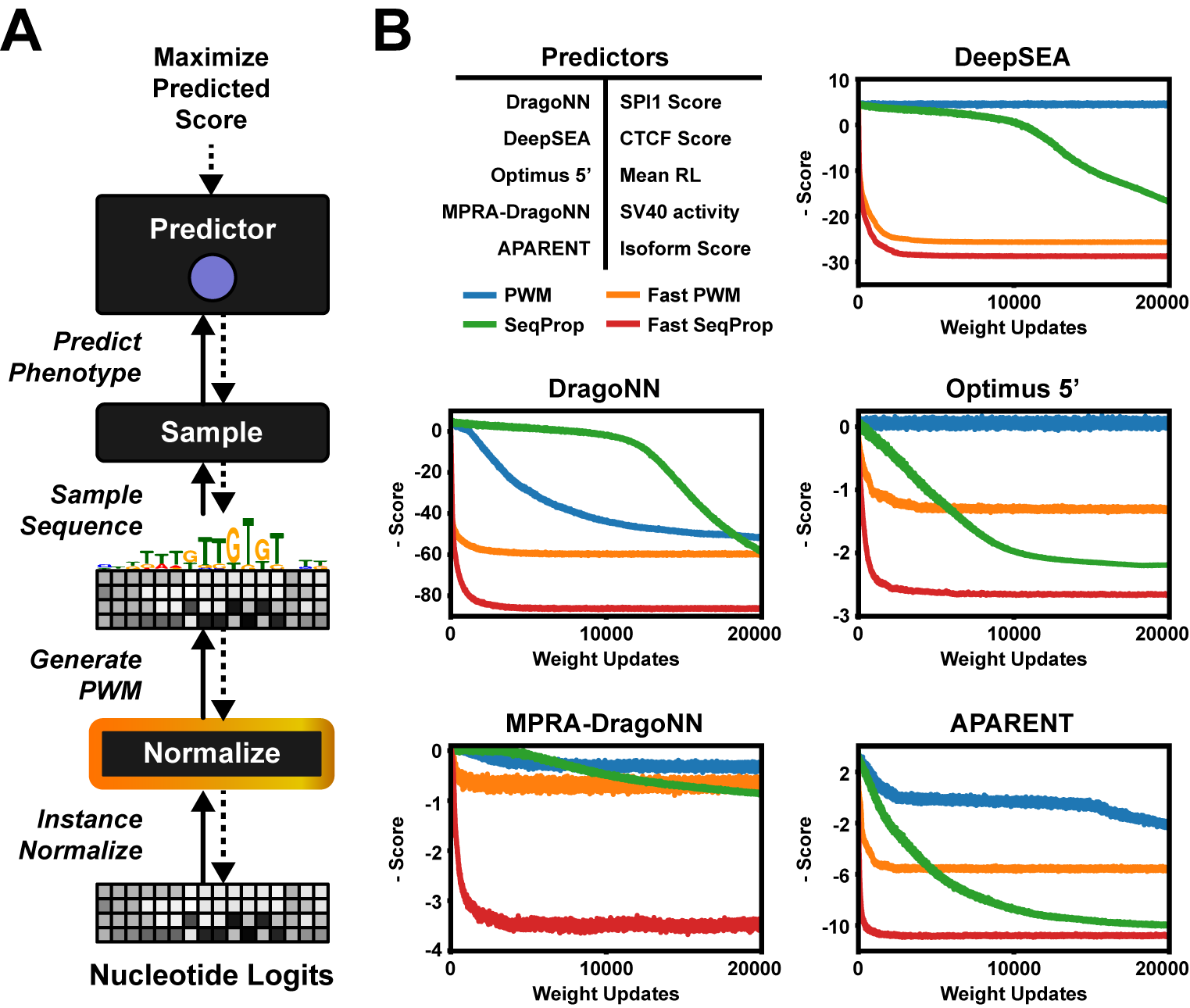}
  \caption{(A) The Fast SeqProp pipeline. A normalization layer is prepended to a softmax layer, which is used as parameters to a sampling layer. (B) Maximizing the predictors DragoNN (SPI1), DeepSEA (CTCF Dnd41), MPRA-DragoNN (SV40), Optimus 5' and APARENT. }
  \label{fig:1}
\end{figure}

Inspired by instance normalization in image GANs (Ulyanov et al., 2016), we hypothesized that the main bottleneck in earlier design methods --  both in terms of optimization speed and minima found -- stem from overly large and disproportionally scaled nucleotide logits. Here, we mitigate this problem by normalizing the logits across positions, i.e. we insert a normalization layer between the trainable logits $\bm{l}_{ij}$ and the sampling layer $\delta(\bm{l})_{ij}$ (Figure \ref{fig:1}A).

For DNA sequence design, where the number of one-hot channels $M$ is small ($M = 4$), we use a normalization scheme commonly referred to as \textit{instance}-normalization. In this scheme, the nucleotide logits of each channel are normalized independently across positions. Let $\bar{\mu}_{j} = \frac{1}{N} \sum_{i=1}^{N} \bm{l}_{ij}$ and $\bar{\varepsilon}_{j} = \sqrt{\frac{1}{N} \sum_{i=1}^{N} (\bm{l}_{ij} - \bar{\mu}_{j})^2}$ be the sample mean and deviation of logits for nucleotide $j$ across all positions $i$. For each step of gradient ascent, we compute the normalized logits $\bm{l}_{ij}^{(\text{norm})}$ as:
\begin{align} \label{eq:7}
    \bm{l}_{ij}^{(\text{norm})} = \frac{ \bm{l}_{ij} - \bar{\mu}_{j} }{ \bar{\varepsilon}_{j}^2 }
\end{align}

Since logits with zero mean and unit variance have limited expressiveness when used as parameters to a probability distribution, we associate each channel $j$ with a global scaling parameter $\gamma_{j}$ and offset $\beta_{j}$. Having an independent offset $\beta_{j}$ per channel is particularly well-suited for DNA, as nucleotides are often associated with a global preferential bias. The scaled, re-centered logits are calculated as:
\begin{align} \label{eq:8}
    \bm{l}_{ij}^{(\text{scaled})} = \bm{l}_{ij}^{(\text{norm})} * \gamma_{j} + \beta_{j}
\end{align}

For protein sequence design, the number of one-hot channels $M$ is considerably larger ($M = 20$), resulting in fewer samples per channel and noisier normalization statistics. Here we found that \textit{layer}-normalization was more stable: We compute a global mean $\bar{\mu} = \frac{1}{N \cdot M} \sum_{i=1}^{N} \sum_{j=1}^{M} \bm{l}_{ij}$ and deviation $\bar{\varepsilon} = \sqrt{\frac{1}{NM} \sum_{i=1}^{N} \sum_{j=1}^{M} (\bm{l}_{ij} - \bar{\mu}_{j})^2}$, and use a shared scale $\gamma$ and offset $\beta$ for all $M$ channels.

Given the normalized and scaled logits $\bm{l}^{(\text{scaled})}$ as parameters for the nucleotide sampler $\delta$ defined in Equation \ref{eq:5}, we maximize $\mathcal{P}(\delta(\bm{l}^{(\text{scaled})}))$ with respect to $\bm{l}_{ij}$, $\gamma_{j}$ and $\beta_{j}$ (or $\gamma$ and $\beta$ in the context of proteins). Using the softmax ST estimator from Equation \ref{eq:6}, we arrive at the following gradients:
\begin{align} \label{eq:9}
    \frac{\partial \mathcal{P}(\delta(\bm{l}^{(\text{scaled})}))}{\partial \bm{l}_{ij}} &= \sum_{k=1}^{M} \frac{\partial \mathcal{P}(\delta(\bm{l}^{(\text{scaled})}))}{\partial \delta(\bm{l}^{(\text{scaled})})_{ik}} \cdot \frac{\partial \sigma(\bm{l}^{(\text{scaled})})_{ik}}{\partial \bm{l}_{ij}^{(\text{scaled})}} \cdot \gamma_{j}\\ \label{eq:10}
    \frac{\partial \mathcal{P}(\delta(\bm{l}^{(\text{scaled})}))}{\partial \gamma_{j}} &= \sum_{i=1}^{N} \sum_{k=1}^{M} \frac{\partial \mathcal{P}(\delta(\bm{l}^{(\text{scaled})}))}{\partial \delta(\bm{l}^{(\text{scaled})})_{ik}} \cdot \frac{\partial \sigma(\bm{l}^{(\text{scaled})})_{ik}}{\partial \bm{l}_{ij}^{(\text{scaled})}} \cdot \bm{l}_{ij}^{(\text{norm})}\\ \label{eq:11}
    \frac{\partial \mathcal{P}(\delta(\bm{l}^{(\text{scaled})}))}{\partial \beta{j}} &= \sum_{i=1}^{N} \sum_{k=1}^{M} \frac{\partial \mathcal{P}(\delta(\bm{l}^{(\text{scaled})}))}{\partial \delta(\bm{l}^{(\text{scaled})})_{ik}} \cdot \frac{\partial \sigma(\bm{l}^{(\text{scaled})})_{ik}}{\partial \bm{l}_{ij}^{(\text{scaled})}}
\end{align}

The normalization removes logit drift by keeping the values proportionally scaled and centered at zero ($\mathrm{E}[l_{ij}^{(\text{norm})}] = 0$, $\mathrm{Var}[l_{ij}^{(\text{norm})}] = 1$), enabling the gradients to swap nucleotides with few updates. Furthermore, the scaling parameter $\gamma_{j}$ adaptively adjusts the sampling entropy to control global versus local optimization. This can be deduced from the gradient components of $\gamma_{j}$ in Equation \ref{eq:10}:
\begin{enumerate}
    \item $\frac{\partial \mathcal{P}(\delta(\bm{l}^{(\text{scaled})}))}{\partial \delta(\bm{l}^{(\text{scaled})})_{ik}}$ is positive for nucleotides which increase fitness and negative otherwise.
    \item $\frac{\partial \sigma(\bm{l}^{(\text{scaled})})_{ik}}{\partial \bm{l}_{ij}^{(\text{scaled})}}$ is positive when $j=k$ and negative otherwise.
    \item $\bm{l}_{ik}^{(\text{norm})}$ is positive only when we are likely to sample the corresponding nucleotide.
\end{enumerate}
Here, the product of the first two terms, $\frac{\partial \mathcal{P}(\delta(\bm{l}^{(\text{scaled})}))}{\partial \delta(\bm{l}^{(\text{scaled})})_{ik}} \cdot \frac{\partial \sigma(\bm{l}^{(\text{scaled})})_{ik}}{\partial \bm{l}_{ij}^{(\text{scaled})}}$, is positive if $j = k$ and nucleotide $j$ raises fitness or if $j \neq k$ and nucleotide $k$ lowers fitness. Put together, the gradient for $\gamma_{j}$ increases when our confidence $\bm{l}_{ij}^{(\text{norm})}$ in nucleotide $j$ is \textit{consistent} with its impact on fitness, such that $\text{sign}\left(\sum_{k=1}^{M} \frac{\partial \mathcal{P}(\delta(\bm{l}^{(\text{scaled})}))}{\partial \delta(\bm{l}^{(\text{scaled})})_{ik}} \cdot \frac{\partial \sigma(\bm{l}^{(\text{scaled})})_{ik}}{\partial \bm{l}_{ij}^{(\text{scaled})}}\right) = \text{sign}\left(\bm{l}_{ij}^{(\text{norm})}\right)$. Conversely, \textit{inconsistent} nucleotides decrement the gradient. At the start of optimization, $\gamma_{j}$ is small, leading to high PWM entropy and large jumps in sequence design space. As we sample consistent nucleotides and the entropy gradient $\frac{\partial \mathcal{P}(\delta(\bm{l}^{(\text{scaled})}))}{\partial \gamma_{j}}$ turns positive, $\gamma_{j}$ increases. Larger $\gamma_{j}$ lowers the entropy and leads to more localized optimization. However, if we sample sufficiently many inconsistent nucleotides, the gradient of $\gamma_{j}$ may turn negative, again raising entropy and promoting global exploration.

Note that, in the context of protein design where we have a single scale $\gamma$ and offset $\beta$, the gradient expressions from Equation \ref{eq:10} and \ref{eq:11} are additively pooled across all $M$ channels. The argued benefits of instance-normalization above thus holds true for layer-normalization as well.

\section{Results}

\subsection{Maximizing Nucleic Acid Sequence-Predictive Neural Networks}

We first evaluate our method on the task of maximizing the classification or regression scores of five DNA- or RNA-level deep neural network predictors: (1) \textit{DragoNN}, a model trained on ChIP-seq data to predict Transcription Factor (TF) binding (in this case binding of SPI1), (2) \textit{DeepSEA} (Zhou et al., 2015), which predicts multiple TF binding probabilities and chromatin modifications (we use it here to maximize the probability of CTCF binding in the cell type Dnd41), (3) \textit{APARENT} (Bogard et al., 2019), which predicts alternative polyadenylation isoform abundance given an input polyadenylation signal, (4) MPRA-DragoNN (Movva et al., 2019), a neural network trained to predict transcriptional activity of short enhancer sequences and, finally, (5) \textit{Optimus 5'} (Sample et al., 2019), which predicts ribosomal load (translational efficiency) of 5' UTR sequences. 

We compare our new logit-normalized, straight-through sampled sequence design method (\textit{Fast SeqProp}) to the previous versions of the algorithm, namely  the original method with continuous softmax-relaxed inputs (Killoran et al., 2017, here referred to as \textit{PWM}) and \textit{SeqProp}, the categorical sampling method described in (Bogard et al., 2019) using the (non-normalized) softmax straight-through gradient estimator. We also tested a logit-normalized version of the softmax-relaxed method, \textit{Fast PWM}, in order to disentangle the individual performance contributions of the normalization scheme and the sampling scheme with respect to the baseline PWM method.

Figure \ref{fig:1}B shows the result of using the design methods to generate maximally scored sequences for each of the five DNA-based predictors. Fast SeqProp converges to $95\%$ - $99\%$ of its minimum test loss within $2,000$ logit updates, and reaches $50\%$ of the minimum loss after only $200$ updates for all predictors except MPRA-DragoNN and Optimus 5'. In contrast, PWM and SeqProp do not converge within $20,000$ updates. 
Fast SeqProp converges to up to 3-fold better optima than all other compared methods. In fact, Fast SeqProp reaches the same or better optima in $200$ updates than the competing methods reach in $20,000$ updates for DragoNN, MPRA-DragoNN and DeepSEA, marking a 100x speedup. For Optimus 5' and APARENT, the speedup is 20x-50x. In addition to gradient-based methods, we demonstrate improved performance compared to discrete search algorithms such as Simulated Annealing (see Supplementary Information, Figure S1A-B). 

In the Supplementary Information, we provide additional technical comparisons of Fast SeqProp to previous activation maximization methods. For example, in Figure S1C, we demonstrate that certain sequence-predictive neural networks suffer from out-of-distribution (OOD) pathologies on continuous sequence relaxations as input, explaining the poor performance of the PWM design method. We further show that adding an entropy penalty to the PWM method still cannot close the performance gap to Fast SeqProp (Figure S1D) and that the Softmax ST estimator is superior to Gumbel Sampling (Figure S1E). Finally, we show that Fast SeqProp appears robust to the choice of optimizer parameters (Figure S1F) and that the Softmax ST estimator outperforms the original ST estimator (Figure S1G).

\subsection{Recapitulating \textit{cis}-Regulatory Biology with Activation Maximization}

\begin{figure}[H]
  \centering
  \includegraphics[scale=0.9]{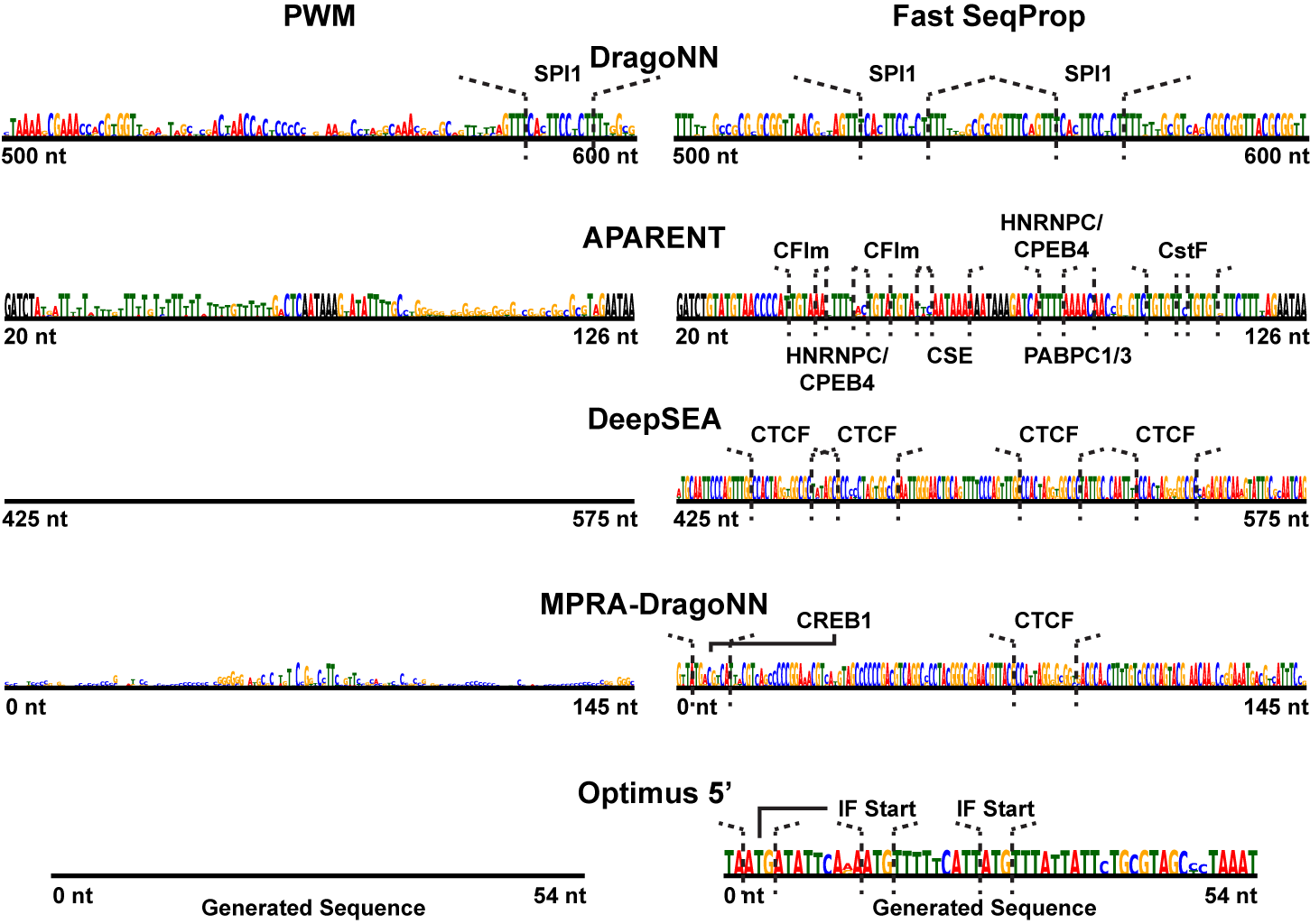}
  \caption{ Example softmax sequences (PSSMs) generated by the PWM and Fast SeqProp methods after 20,000 updates of gradient ascent updates with default optimizer parameters (Adam). The logit matrices $\bm{l}$ were uniformly randomly initialized prior to optimization. Identified cis-regulatory motifs annotated above each sequence. }
  \label{fig:2}
\end{figure}

In Figure \ref{fig:2} we compare example sequence optimizations of the PWM and Fast SeqProp methods. As can be seen, even after $20,000$ updates, the PWM method has not converged for most of the tested predictors. In contrast, we find plenty of \textit{cis}-regulatory motifs in the converged sequences generated by Fast SeqProp. Since our method was tasked with \textit{maximizing} the predicted score of each model, we would expect to find enhancing motifs and regulatory logic embedded in the sequences which give rise to these extreme model responses.

For example, when maximizing DragoNN, Fast SeqProp generates dual SPI1 binding motifs (Sandelin et al., 2004). For APARENT, Fast SeqProp generates multiple CFIm binding motifs, dual CSE hexamers, and multiple cut sites with CstF binding sites. These are all regulatory motifs known to enhance cleavage and polyadenylation by stimulating recruitment of the polyadenylation machinery (Di Giammartino et al., 2011; Shi et al., 2012; Elkon et al., 2013; Tian \& Manley, 2017). For DeepSEA, Fast SeqProp generates four CTCF binding sites. For MPRA-DragoNN, we identify both CRE- and a CTCF binding sites embedded within a GC-rich context, which aligns well with what we might expect to find in a strong enhancer  (Kheradpour et al., 2014; Ernst et al., 2016). Finally, for Optimus 5’, Fast SeqProp generates a T-rich sequence with multiple in-frame (IF) uAUGs. These determinants were found to improve ribosome loading  (Sample et al., 2019). See the Supplementary Information (Figure S2) for additional visualizations comparing the PWM and Fast SeqProp methods during different stages of optimization.

\subsection{Regularized Sequence Design}

\begin{figure}[H]
  \centering
  \includegraphics[scale=0.75]{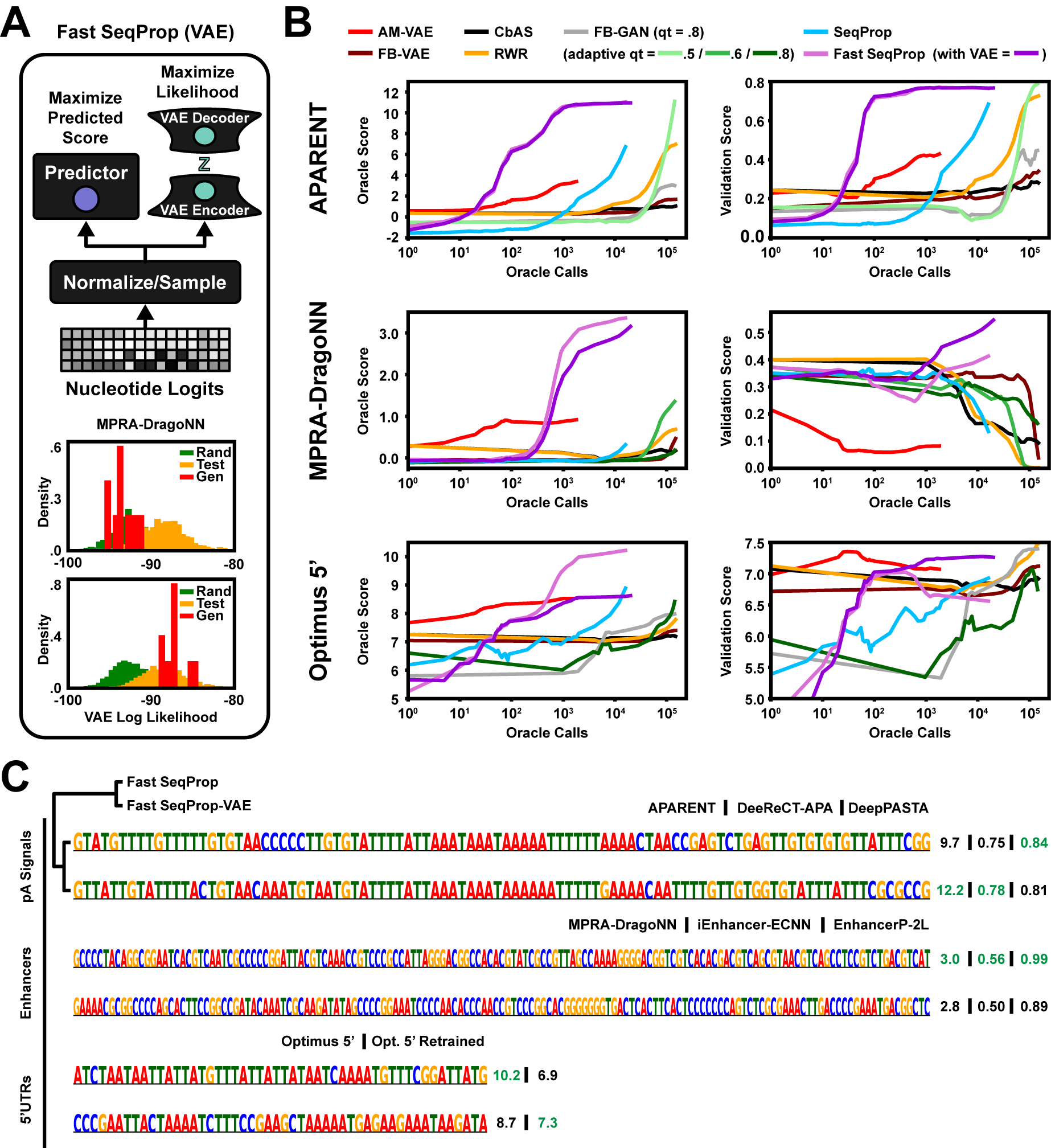}
  \caption{ (A) Top: VAE-regularized Fast SeqProp. A variational autoencoder (VAE) is used to control the estimated likelihood of designed sequences during gradient ascent optimization. Bottom: Estimated VAE log likelihood distribution of random sequences (green), test sequences from the MPRA-DragoNN dataset (orange) and designed sequences (red), using Fast SeqProp without and with VAE regularization (top and bottom histogram respectively). (B) Oracle fitness score trajectories (APARENT, MPRA-DragoNN and Optimus 5') and validation model score trajectories (DeeReCT-APA, iEnhancer-2L and retrained Optimus 5') as a function of the cumulative number of predictor calls made during the sequence design phase. Shown are the median scores across 10 samples per design method, for three repeats. (C) Example designed sequences for APARENT, MPRA-DragoNN and Optimus 5', using Fast SeqProp with and without VAE-regularization. Oracle and validation model scores are annotated on the right. }
  \label{fig:3}
\end{figure}

While finding regulatory logic in the sequences produced by activation maximization is a good indication that we actually generate patterns with biological meaning, the method may still not be suitable in its most basic form for sequence design. There is the potential issue of overfitting to the predictor oracle during sequence optimization, as the oracle may lose its accuracy when drifting out of the training data distribution to maximize predicted fitness. By training a differentiable likelihood model, such as a variational autoencoder (VAE; Kingma et al. 2013), on samples from the same data  and using it as a regularizer in the cost function, we can prevent drift to low-confidence regions of design space (Figure \ref{fig:3}A; top). Using a  VAE to avoid drift has previously been demonstrated by (Gómez-Bombarelli et al. 2018; Brookes et al., 2019; Linder et al. 2020). In summary, we extend the original optimization objective (Eq. \ref{eq:1}) by passing the sampled one-hot pattern $\delta(\bm{l})$ to the VAE and penalize the pattern based on its VAE-estimated marginal likelihood, $p_{\text{VAE}}(\delta(\bm{l}))$, using ST approximation for backpropagation (see Equation \ref{eq:vae-seqprop} in Methods).

The degree to which predictors exhibit pathological behavior when maximized seems to vary on a case-by-case basis and likely depends heavily on the data distribution. When designing maximally strong gene enhancers using the MPRA-DragoNN predictor, for example, VAE-regularization has a clear effect on shifting the distribution of the designed sequences (Figure \ref{fig:3}A; bottom histograms).
In contrast, when designing polyadenylation signals, VAE-regularization has no effect since non-regularized optimization already generates sequences that are at least as likely as training data according to the VAE (see Supplementary Information, Figure S3A).

Next, we tasked the VAE-regularized Fast SeqProp method with designing maximally strong polyadenylation signals (using APARENT as the oracle), maximally transcriptionally active enhancer sequences (using MPRA-DragoNN as the oracle) and maximally translationally efficient 5' UTRs (using Optimus 5'). For each task, we trained a $\beta$-VAE and W-GAN on a sample of $5,000$ high-fitness sequences (see Methods for details). We then used the methods CbAS (Brookes et al., 2019), FB-GAN (Gupta et al., 2019), AM-VAE (Killoran et al., 2017), RWR (Peters et al., 2007) and FB-VAE (VAE-version of FB-GAN) to maximize each oracle, using the VAE or GAN we trained earlier and default method parameters. We used the same VAE as the regularizer for our design method (Fast SeqProp). During optimization, we measured the fitness scores of both the oracle and a number of independent validation models that we did not directly optimize for, allowing us to estimate sequence fitness in an unbiased way. Specifically, when designing polyadenylation signals based on APARENT, we validated the designs using DeeReCT-APA (Li et al., 2020), an LSTM trained on 3'-sequencing data of mouse cells, and DeepPASTA (Arefeen et al., 2019), a CNN trained on 3'-sequencing data of multiple human tissues (we used the model trained on brain tissue). When designing enhancer sequences, we validated the designs using iEnhancer-ECNN (Nguyen et al., 2019), an ensemble of CNNs trained on genomic enhancer sequences, and EnhancerP-2L (Butt et al., 2020), a Random Forest-classifier based on statistical features extracted from enhancer regions in the genome. Finally, to validate Optimus 5' designs, we had access to a newer version of the model that had been trained on additional MPRA data, making it more robust particularly on outlier sequences such as long homopolymer stretches (Sample et al, 2019). On a practical note, we found it quite difficult to train a VAE on the APARENT, Optimus 5' and MPRA-DragoNN datasets, and the convergence of CbAS, RWR and FB-GAN appeared sensitive to quantile threshold settings, which we believe stem from the considerable data heterogeneity and variability.

The results (Figure \ref{fig:3}B) show that Fast SeqProp reaches significantly higher oracle fitness scores and validation model scores with orders of magnitudes fewer calls to the oracle for all tasks except the 5' UTR design problem, where instead AM-VAE reaches high validation scores faster. The other methods either do not reach the same median validation score in the total allotted time, or do so at the expense of reduced diversity (see Supplementary Information, Figure S3B). For the polyadenylation signal design task, Fast SeqProp reaches identical DeeReCT-APA (Figure \ref{fig:3}B; top right) and DeepPASTA (see Supplementary Information, Figure S3C) validation scores with or without VAE-regularization. The designed polyadenylation signal sequences look quite similar and share many motifs, such as multiple CFIm, CstF and CPSF binding sites (Figure \ref{fig:3}C; top). For the enhancer design task, the VAE-regularization is clearly beneficial according to the validation model; while enhancers designed by Fast SeqProp without the VAE have a median MPRA-DragoNN score of $3.5$, the median iEnhancer-ECNN score (Figure \ref{fig:3}B; middle right) is just $0.43$. With VAE-regularization, we generate sequences with a lower median MPRA-DragoNN score ($3.25$), but higher iEnhancer-ECNN score ($0.55$). However, closer inspection reveals that Fast SeqProp does not consistently generate worse enhancers according to the validation model than its VAE-regularized counterpart. Rather, Fast SeqProp without VAE either generates highly scored enhancers by the validation model or lowly scored sequences, while Fast SeqProp with VAE consistently generates medium-scored enhancers (example sequences are shown in Figure \ref{fig:3}C; middle). This dynamic is also observed with another validation model (EnhancerP-2L; see Supplementary Information, Figure S3D); Only $80\%$ of Fast SeqProp (no VAE) sequences are identified by EnhancerP-2L as enhancers, while nearly $100\%$ of Fast SeqProp-VAE sequences are identified. However, their weighted predicted enhancer strengths are identical. It is also worth noting that most other methods decrease their validation scores when increasing their MPRA-DragoNN scores; this is because they get stuck in a suboptimal, local minimum with pathological AT-repeats. Finally, VAE-regularization is beneficial for designing 5' UTRs, as it restricts the sequences from becoming overly T-rich, a sequence pathology present in the original Optimus 5' model which the retrained version understands actually decreases ribosome load (Figure \ref{fig:3}B; bottom; Figure \ref{fig:3}C; bottom).

In the Supplementary Information, we provide extra benchmark experiments comparing Fast SeqProp to a subset of the above design methods. In particular, in Figure S3E, we train the same kind of oracles as was used by Brookes et al. (2019) to estimate uncertainty in the fitness predictions (Lakshminarayanan et al. 2017), and use these models to replicate the polyadenylation signal and 5' UTR design benchmarks. We also replicate the GFP design task used in Brookes et al. (2019). Additionally, in Figure S3F, we include an example where we use MPRA-DragoNN to design maximally specific enhancers in the cell line HepG2 (and inactivated in K562), and show how internal network penalties can be used to regularize the sequence optimization when it is hard to train an uncertainty-estimator oracle that is sufficiently accurate.

\subsection{Protein Structure Optimization}

\begin{figure}[H]
  \centering
  \includegraphics[scale=0.85]{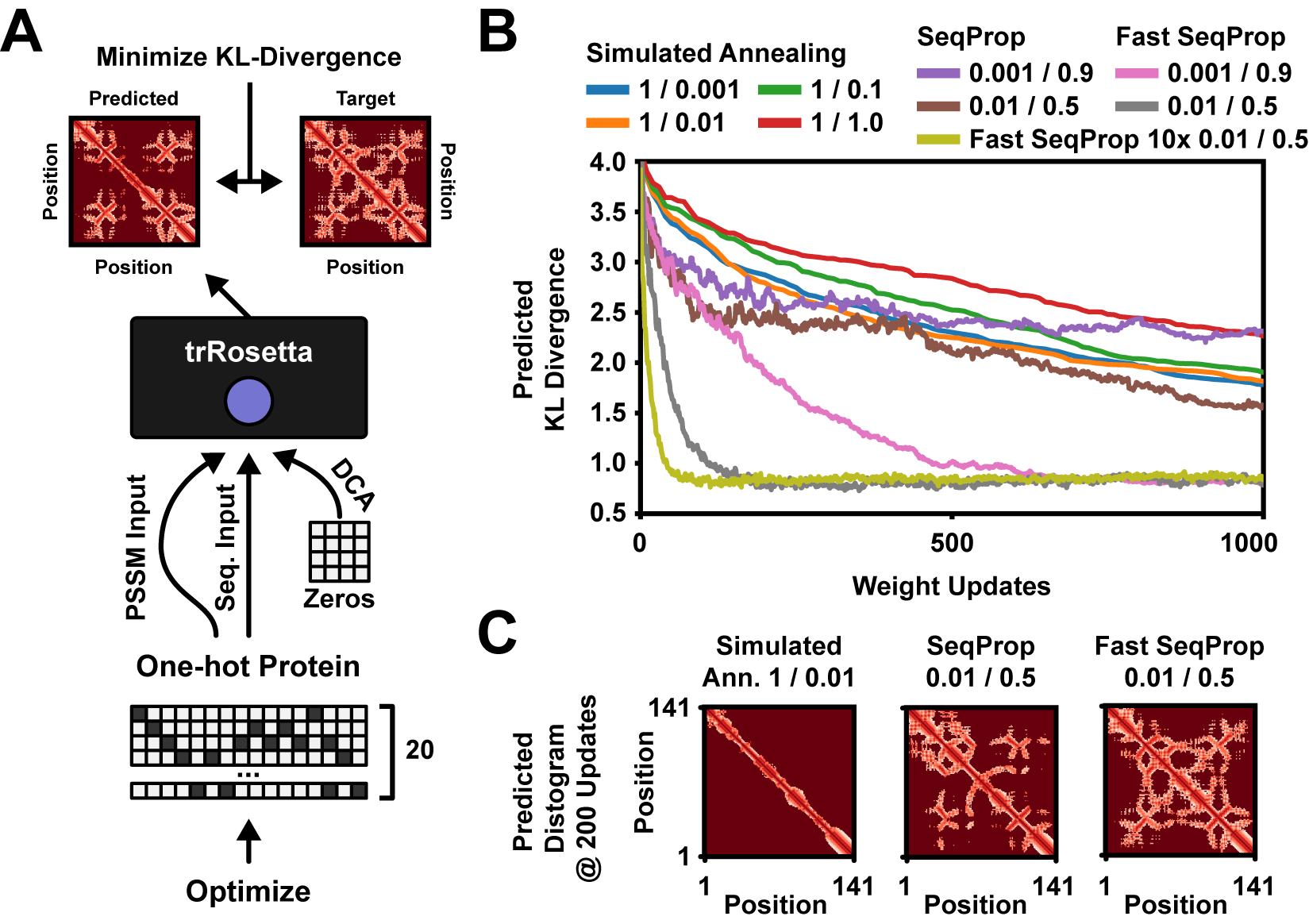}
  \caption{ (A) Protein sequences are designed to minimize the KL-divergence between predicted and target distance and angle distributions. The one-hot pattern is used for two of the trRosetta inputs. (B) Generating sequences which conform to the target predicted structure of a Sensor Histidine Kinase. Simulated Annealing was tested at several initial temperatures, with 1 substitution per step. Similarly, SeqProp and Fast SeqProp was tested at several combinations of learning rate and momentum. (C) Predicted residue distance distributions after 200 iterations. }
  \label{fig:4}
\end{figure}

\noindent Multiple deep learning models have recently been developed for predicting tertiary protein structure (AlQuraishi, 2019; Senior et al., 2020; Yang et al., 2020). Here, we demonstrate our method by designing \textit{de novo} protein sequences which conform to a target residue contact map as predicted by trRosetta (Yang et al., 2020). The predictor takes three inputs (Figure \ref{fig:4}A): A one-hot coded sequence, a PSSM constructed from a multiple-sequence alignment (MSA) and a direct-coupling analysis (DCA) map. For our design task, we pass the optimizable one-hot pattern to the first two inputs and an all-zeros tensor as the DCA feature map. Given the predicted distance distribution $\bm{D}^{P} \in [0, 1]^{N \times N \times 37}$ and angle distributions $\bm{\theta}^{P}, \bm{\omega}^{P} \in [0, 1]^{N \times N \times 24}$, $\bm{\phi}^{P} \in [0, 1]^{N \times N \times 12}$, we minimize the mean KL-divergence against target distributions $\bm{D}^{T}$, $\bm{\theta}^{T}$, $\bm{\omega}^{T}$ and $\bm{\phi}^{T}$:
\begin{equation}
\begin{aligned}
    \min_{\bm{l}} & \quad \text{KL}(\bm{D}^{P} || \bm{D}^{T}) + \text{KL}(\bm{\theta}^{P} || \bm{\theta}^{T}) + \text{KL}(\bm{\omega}^{P} || \bm{\omega}^{T}) + \text{KL}(\bm{\phi}^{P} || \bm{\phi}^{T})\\
    \text{where} & \quad \text{KL}(\bm{X} || \bm{Y}) = \frac{1}{N^{2}} \cdot \sum_{i=1}^{N} \sum_{j=1}^{N} \sum_{k=1}^{K} \bm{Y}_{ijk} \cdot \log \bigg( \frac{\bm{Y}_{ijk}}{\bm{X}_{ijk}} \bigg)
\end{aligned}
\end{equation}

\noindent We compared SeqProp and Simulated Annealing to a modified version of Fast SeqProp, where logits are normalized across all residue channels (\textit{layer}-normalized rather than \textit{instance}-normalized) to reduce the increased variance of $20$ one-hot channels. We used the methods to design protein sequences which conformed to the target structure of an example protein (Sensor Histidine Kinase). We optimized $5$ independent sequences per design method and recorded the median KL-loss at each iteration. The results show that Fast SeqProp converges faster and reaches better minima (Figure \ref{fig:4}B; see also Supplementary Information, Figure S4A); after 200 iterations, Fast SeqProp reached 4x lower KL-divergence than all other methods, and much of the target structure is visible (Figure \ref{fig:4}C). While the choice of learning rate changes the rate of convergence, it does not alter the minima found by Fast SeqProp (Figure \ref{fig:4}B). Additionally, by sampling multiple sequences at once and walking down the average gradient (e.g. 10 samples per update), we can improve the rate of convergence further by making the gradient less noisy (see Supplementary Information, Figure S4B). Importantly, this scales significantly better than linear in execution time, since multiple samples can be differentiated in parallel on a GPU. Finally, we replicated our results by designing sequences for a different protein structure (an alpha-helical hairpin; see Supplementary Information Figure S4C-E).

\section{Discussion}

By normalizing nucleotide logits across positions and using a global entropy parameter, \textit{Fast SeqProp} keep logits proportionally scaled and centered at zero. The gradient of the entropy parameter $\gamma$ in our design method adaptively adjusts the sampling temperature to trade off global and local optimization. In the beginning, $\gamma$ is small, corresponding to a high PWM entropy and consequently very diverse sequence samples. As optimization progresses, $\gamma$ grows, leading to more localized sequence changes. This adaptive mechanism, in combination with flexible nucleotide logits due to the normalization, results in a highly efficient design method. As demonstrated on five deep learning predictors, logit normalization enables extremely fast sequence optimization, with a 50-to-100-fold speedup compared to state-of-the-art methods for many predictors, and with better predicted optima.  Highlighting its broad usefulness, since this paper was first deposited on a pre-print server, several genomic and protein design papers based on variations of normalized and discretized activation maximization have been made available (Schreiber et al., 2020; Norn et al., 2020; Tischer et al., 2020).

In addition to logit drift and vanishing gradients, the original sequence ascent method suffers from predictor pathologies due to passing continuous softmax sequence relaxations as input, a problem fully removed by using discrete sampling. We further observed that straight-through sampling leads to consistently better optima than softmax relaxation, suggesting that it traverses local minima. In fact, our method outperformed global optimization meta heuristics such as Simulated Annealing on more difficult design tasks, such as designing 1000 nt long enhancer regions or designing protein sequences which conform to a complex target structure. We further demonstrated robust sequence design results on a number of tasks, including the design of strong alternative polyadenylation signals, efficiently translated 5' UTRs and enhancers that result in high transcriptional activity. We did so by incorporating a regularization penalty based on variational autoencoders and showed better and faster convergence than other regularized design methods.

\section{Conclusion}

We presented an improved version of gradient ascent (or activation maximization) for biological sequence design, called \textit{Fast SeqProp}, which combines logit normalization with stochastic nucleotide sampling and straight-through gradients. We demonstrated the efficacy of the method on several DNA-, RNA- and protein design tasks. We expect this algorithmic improvement to be broadly useful to the research community for biomolecular optimization at the level of primary sequence. The approach introduced here could accelerate the design of functional enzymes and other biomolecules, potentially resulting in novel drug therapies, molecular sensors and more.



\section*{Methods}

\subsection*{Activation Maximization Design Methods}

In Figure \ref{fig:1} and throughout the paper, we compare four different activation maximization methods for sequences: (1) \textit{Fast SeqProp} (Our method) -- The modified activation maximization method which combines the logit normalization scheme of Equation \ref{eq:7}-\ref{eq:8} with the softmax straight-through estimator of Equation \ref{eq:5}-\ref{eq:6}, (2) \textit{PWM} -- The original method with continuous softmax-relaxed inputs (Killoran et al., 2017), (3) \textit{SeqProp} -- The categorical sampling method described in (Bogard et al., 2019) using the (non-normalized) softmax straight-through gradient estimator, and (4) \textit{Fast PWM} -- A logit-normalized version of the softmax-relaxed method.

Starting with a randomly initialized logit matrix $\bm{l}$, for the methods PWM and Fast PWM we optimize $\bm{l}$ using the softmax relaxation $\sigma(\bm{l})$ from Equation \ref{eq:3}. For SeqProp and Fast SeqProp, we optimize $\bm{l}$ using the discrete nucleotide sampler $\delta(\bm{l})$ from Equation \ref{eq:5}. We define the optimization loss (or the 'train' loss) as:
\begin{align*}
    \mathcal{L}_{\text{train}}(\bm{l}) = - \mathcal{P}(\bm{x}(\bm{l}))
\end{align*}
For PWM and Fast PWM, $\bm{x}(\bm{l}) = \sigma(\bm{l})$. For SeqProp and Fast SeqProp, $\bm{x}(\bm{l}) = \delta(\bm{l})$.

The actual loss (or 'test' loss) is evaluated on the basis of discrete sequence samples drawn from the optimized softmax representation $\sigma(\bm{l})$, regardless of design method. In all four methods, we can use the categorical nucleotide sampler $\delta(\bm{l})$ to draw sequence samples and compute the mean test loss as:
\begin{align*}
	 \mathcal{L}_{\text{test}}(\{\bm{l}^{(k)}\}_{k=1}^{K}) = - \frac{1}{K} \frac{1}{S} \sum_{k=1}^{K} \sum_{s=1}^{S} \mathcal{P}(\delta(\bm{l}^{(k)})^{(s)})
\end{align*}
Here $S$ refers to the number of samples drawn from each softmax sequence $\sigma(\bm{l}^{(k)})$ at every weight update $t$, and $K$ is the number of independent optimization runs. In all experiments, we set $K = 10$ and $S = 10$.

In addition to gradient-based methods, we compare Fast SeqProp to discrete search algorithms. The first method is a pairwise nucleotide-swapping search (\textit{Evolution}; Sample et al., 2019), where sequence $\bm{x}$ is mutated with either $1$ or, with a 50\% chance, $2$ random substitutions at each iteration, resulting in a new candidate sequence $\bm{x}'$. $\bm{x}'$ is only accepted if $\mathcal{P}(\bm{x}') > \mathcal{P}(\bm{x})$. We also tested a well-known meta heuristic -- \textit{Simulated Annealing} (Kirkpatrick et al., 1983). In Simulated Annealing, mutations are accepted even if they result in lower fitness with probability $P(\bm{x}', \bm{x}, T)$, where $T$ is a temperature parameter. Here we use the Metropolis acceptance criterion (Metropolis et al., 1953):
\begin{align*}
    P(\bm{x}', \bm{x}, T) = e^{-(\mathcal{P}(\bm{x}) - \mathcal{P}(\bm{x}')) / T}
\end{align*}

\subsubsection*{VAE-regularized Fast SeqProp}

In the main paper (Figure \ref{fig:3}), we use a variational autoencoder (VAE; Kingma et al. 2013) to regularize the sequence design when running Fast SeqProp. The original optimization objective (Eq. \ref{eq:1}) is extended by passing the sampled one-hot pattern $\delta(\bm{l})$ to the VAE and estimating its marginal likelihood, $p_{\text{VAE}}(\delta(\bm{l}))$, using importance-weighted inference. We then minimize a margin loss with respect to the mean likelihood $p_{\text{ref}}$ of the original training data to keep sequence designs in-distribution, using the Softmax ST estimator to propagate gradients back to $\bm{l}$:
\begin{align} \label{eq:vae-seqprop}
    \min_{\bm{l}} - \mathcal{P}(\delta(\bm{l})) + \lambda \cdot \text{max}\big[\log_{10} p_{\text{ref}} - \log_{10} p_{\text{VAE}}(\delta(\bm{l})) - \rho , 0 \big]
\end{align}

\subsubsection*{VAE-regularized Fast SeqProp with Uncertainty-estimation}

In the Supplementary Information (Figure S3E), we replicate the benchmark comparison of the main paper (Figure \ref{fig:3}), but we use oracle predictors capable of estimating the uncertainty in their fitness predictions to further regularize the designs (Lakshminarayanan et al. 2017). Assume that the oracle model predicts the mean $\mu\big[\delta(\bm{l})\big]$ and standard deviation $\epsilon\big[\delta(\bm{l})\big]$ of fitness scores for the designed (sampled) pattern $\delta(\bm{l})$. We then use the (differentiable) survival function of the normal distribution to maximize the probability $p_{\mu[\delta(\bm{l})], \epsilon[\delta(\bm{l})]}(\mathbbm{Y} > q)$ that the predicted fitness of sequence $\delta(\bm{l})$ is larger than quantile $q$ of the training data:
\begin{align} \label{eq:balaji-vae-seqprop}
    \min_{\bm{l}} - \log_{10} p_{\mu[\delta(\bm{l})], \epsilon[\delta(\bm{l})]}(\mathbbm{Y} > q) + \lambda \cdot \text{max}\big[\log p_{\text{ref}} - \log_{10} p_{\text{VAE}}(\delta(\bm{l})) - \rho , 0 \big]
\end{align}

\subsubsection*{VAE-regularized Fast SeqProp with Activity-regularization}

In the Supplementary Information (Figure S3F), we use the predictor MPRA-DragoNN to design maximally HepG2-specific enhancer sequences, and use activity-regularization on (some of) the internal layers of the predictor to regularize the optimization. We maximize the predicted fitness score $\mathcal{P}(\delta(\bm{l}))$ (and minimize the VAE-loss as before) while also minimizing a margin loss applied to the sum of a subset of convolutional activation maps $\mathcal{C}_{k}(\delta(\bm{l}))$:
\begin{equation}
\begin{aligned} \label{eq:act-reg-vae-seqprop}
    \min_{\bm{l}} &- \mathcal{P}(\delta(\bm{l})) + \lambda \cdot \text{max}\big[\log_{10} p_{\text{ref}} - \log_{10} p_{\text{VAE}}(\delta(\bm{l})) - \rho , 0 \big]\\
    &+ \eta_{1} \cdot \text{max}\big[\mathcal{C}_{1}(\delta(\bm{l})) - C_{1}, 0 \big] + ... + \eta_{K} \cdot \text{max}\big[\mathcal{C}_{K}(\delta(\bm{l})) - C_{K}, 0 \big]
\end{aligned}
\end{equation}

\subsection*{Predictor Models}

We designed sequences for five distinct DNA- or RNA deep learning predictors. For each of these models, we defined one of their (potentially many) outputs as the classification or regression score $\mathcal{P}(\bm{x}) \in \mathbb{R}$ to maximize in Equation \ref{eq:1}. We also designed protein sequences according to a 3D protein structure predictor. Here is a brief description of each fitness predictor:\\

\noindent \textbf{DragoNN:} Predicts the probability of SPI1 transcription factor (TF) binding within a 1000-nt sequence. We define $\mathcal{P}(\bm{x})$ as the logit score of the network output. The trained model was downloaded from:\\ {\footnotesize \href{http://mitra.stanford.edu/kundaje/projects/dragonn/SPI1.classification.model.hdf5}{http://mitra.stanford.edu/kundaje/projects/dragonn/SPI1.classification.model.hdf5}}.\\

\noindent \textbf{DeepSEA:} (Zhou et al., 2015) Predicts multiple TF binding probabilities and chromatin modifications in a 1000-nt sequence. We define $\mathcal{P}(\bm{x})$ as the logit score of the CTCF (Dnd41) output. The trained model was downloaded from:\\ {\footnotesize \href{http://deepsea.princeton.edu/media/code/deepsea.v0.94c.tar.gz}{http://deepsea.princeton.edu/media/code/deepsea.v0.94c.tar.gz}}.\\

\noindent \textbf{APARENT:} (Bogard et al., 2019) Predicts proximal alternative polyadenylation isoform abundance in a 206-nt sequence. We define $\mathcal{P}(\bm{x})$ as the logit score of the network output. The trained model was downloaded from:\\ {\footnotesize \href{https://github.com/johli/aparent/tree/master/saved_models}{https://github.com/johli/aparent/tree/master/saved\_models}}.\\

\noindent \textbf{MPRA-DragoNN:} (Movva et al., 2019) Predicts transcriptional activity of a 145-nt promoter sequence. We define $\mathcal{P}(\bm{x})$ as the sixth output (SV40) of the 'Deep Factorized' model. The trained model was downloaded from:\\ {\footnotesize \href{https://github.com/kundajelab/MPRA-DragoNN/tree/master/kipoi/DeepFactorizedModel}{https://github.com/kundajelab/MPRA-DragoNN/tree/master/kipoi/DeepFactorizedModel}}.\\

\noindent \textbf{Optimus 5':} (Sample et al., 2019) Predicts mean ribosome load in a 50-nt sequence. $\mathcal{P}(\bm{x})$ is the (non-scaled) output of the 'evolution' model. The trained model was downloaded from:\\
{\footnotesize \href{https://github.com/pjsample/human_5utr_modeling/tree/master/modeling/saved_models}{https://github.com/pjsample/human\_5utr\_modeling/tree/master/modeling/saved\_models}}.\\

\noindent \textbf{trRosetta:} (Yang et al., 2020) Predicts amino acid residue distance distributions and angle distributions of the input primary sequence. We defined the optimization objective as minimizing the mean KL-divergence between the predicted distance- and angle distributions of the designed sequence compared to a target structure (see the definition in Section 'Protein Structure Optimization' of the main paper). The trained model was downloaded from:\\ {\footnotesize \href{https://files.ipd.uw.edu/pub/trRosetta/model2019\_07.tar.bz2}{https://files.ipd.uw.edu/pub/trRosetta/model2019\_07.tar.bz2}}.\\

\noindent All optimization experiments were carried out in Keras (Chollet, 2015) using Adam with default parameters (Kingma et al., 2014). Some predictor models were ported using \textit{pytorch2keras}.

\subsection*{Validation Models}

When designing sequences for the predictor models listed above, we computed validation scores based on the following held-out models (i.e. models we did not explicitly optimize for):\\

\noindent \textbf{DeeReCT-APA:} (Li et al., 2020) Predicts relative isoform abundances for multiple competing polyadenylation signals. The model was trained on mouse 3' sequencing data. We used the model to score a particular designed polyadenylation signal by predicting its relative use when competing with a strong, fixed distal polyadenylation signal. The model was trained using the code repository at:\\ {\footnotesize \href{https://github.com/lzx325/DeeReCT-APA-repo}{https://github.com/lzx325/DeeReCT-APA-repo}}.\\

\noindent \textbf{DeepPASTA:} (Arefeen et al., 2019) Predicts relative isoform abundance of two competing polyadenylation signals. Several model versions exists, we used the one trained on human brain tissue 3' sequencing data. To score a particular designed polyadenylation signal, we predicted its relative use when competing with a strong distal signal. The trained model was downloaded from:\\ {\footnotesize \href{https://www.cs.ucr.edu/\~aaref001/DeepPASTA\_site.html}{https://www.cs.ucr.edu/\~aaref001/DeepPASTA\_site.html}}.\\

\noindent \textbf{iEnhancer-ECNN:} (Nguyen et al., 2019) Detects genomic enhancer regions and predicts whether it is a weak or strong enhancer. We used the product of these two probability outputs to score each designed enhancer sequence. The model was trained using the code repository at:\\ {\footnotesize \href{https://github.com/ngphubinh/enhancers}{https://github.com/ngphubinh/enhancers}}.\\

\noindent \textbf{EnhancerP-2L:} (Butt et al., 2020) Detects genomic enhancer regions and predicts whether it is a weak or strong enhancer. For a sample of generated sequences per design method, we calculated the mean detect/not detect prediction rate, the mean weak/strong prediction rate and the mean p-score. The model was available via a web application at:\\ {\footnotesize \href{http://biopred.org/enpred/pred}{http://biopred.org/enpred/pred}}.\\

\noindent \textbf{Retrained Optimus 5':} (Sample et al., 2019) A retrained version of Optimus 5', where the training data had been complemented with extreme sequences (such as long single-nucleotide repeats, etc.). The trained model was downloaded from:\\
{\footnotesize \href{https://github.com/pjsample/human_5utr_modeling/tree/master/modeling/saved_models}{https://github.com/pjsample/human\_5utr\_modeling/tree/master/modeling/saved\_models}}.

\subsection*{Auxiliary Models}

In Figure \ref{fig:3}, we trained a variational autoencoder (VAE) and a generative adversarial network (GAN) on a subset of the data that was originally used to train each of the predictor oracles APARENT, MPRA-DragoNN and Optimus 5'. For each design task, we selected a sample of $5,000$ sequences with highest observed fitness and a sample of $5,000$ randomly selected sequences. The VAE, which was based on a residual network architecture (He et al., 2016), was trained on the high-fitness subset of sequences. The W-GAN, which was based on the architecture of Gupta et al., (2019), was trained on the random subset of sequences.

\subsection*{Other Design Methods}

A selection of design methods were used for benchmark comparisons in Figure \ref{fig:3}. Here we describe how they were executed and what parameter settings were used:\\

\noindent \textbf{CbAS:} (Brookes et al., 2019) The procedure was started from the VAE which had been pre-trained on the high-fitness dataset. It was executed for $150$ rounds and, depending on design task, either $100$ or $1,000$ sequences were sampled and used for weighted re-training at the end of each round (whichever resulted in higher fitness scores). The threshold was set to either the $60$th or $80$th pecentile of fitness scores predicted on the training data (whichever resulted in more stable fitness score trajectories). The VAE was trained for either $1$ or $10$ epochs at the end of each round (whichever resulted in more stable fitness scores -- for some tasks, the fitness scores would drop abruptly after only a few sampling rounds when training the VAE for $10$ epochs per round). For the benchmark comparison in the main paper, the standard deviation of the predictions were set to a small constant value ranging between $0.02$ and $0.1$, depending on application (since none of the pre-trained oracles APARENT, MPRA-DragoNN or Optimus 5' predicts deviation, we used a small constant deviation that was $\sim 50$x smaller than the maximum possible predicted value). In the Supplementary Information, where we use oracles with uncertainty estimation, we also supplied the predicted standard deviation to the CbAS survival function. The code was adapted from:\\ {\footnotesize \href{https://github.com/dhbrookes/CbAS/}{https://github.com/dhbrookes/CbAS/}}.\\

\noindent \textbf{RWR:} (Peters et al., 2007) The procedure was started from the VAE which had been pre-trained on the high-fitness dataset. It was executed for $150$ rounds and $100$ or $1000$ sequence samples were used for weighted re-training at the end of each round (whichever resulted in higher fitness scores). The VAE was trained for $10$ epochs each round. The code was adapted from:\\ {\footnotesize \href{https://github.com/dhbrookes/CbAS/}{https://github.com/dhbrookes/CbAS/}}.\\

\noindent \textbf{AM-VAE:} (Killoran et al., 2017) This method performs activation maximization by gradient ascent through a pre-trained VAE in order to design sequences. The procedure was started from the VAE which had been pre-trained on the high-fitness dataset. Each sequence was optimized for $2,000$-$5,000$ updates depending on design task (using the Adam optimizer). A normally distributed noise term was added to the gradients to help overcome potential local minima. The code was adapted from:\\ {\footnotesize \href{https://github.com/dhbrookes/CbAS/}{https://github.com/dhbrookes/CbAS/}}.\\

\noindent \textbf{FB-GAN:} (Gupta et al., 2019) The FB-GAN procedure was started from the W-GAN which had been pre-trained on a random sample of sequences. The method was executed for $150$ epochs and $960$ sequences were sampled and used for feedback at the end of each epoch. We either set the feedback threshold to a fixed value (the $80$th percentile of fitness scores predicted on the high-fitness dataset), or we adaptively re-set the threshold to a certain percentile as measured on the $960$ sampled sequences at the end of each epoch. The code was adapted from:\\ {\footnotesize \href{https://github.com/av1659/fbgan}{https://github.com/av1659/fbgan}}.\\

\noindent \textbf{FB-VAE:} (Gupta et al., 2019) A VAE-based version of the FB-GAN. The procedure was started from the VAE which had been pre-trained on the high-fitness dataset. It was executed for $150$ epochs and $100$ or $1000$ sequence samples were used for feedback at the end of each epoch (whichever resulted in higher fitness scores). A fixed threshold was used (either the $60$th or $80$th percentile as predicted on the high-fitness data). The code was adapted from:\\ {\footnotesize \href{https://github.com/dhbrookes/CbAS/}{https://github.com/dhbrookes/CbAS/}}.


\section*{References}

\small

Alipanahi, B., Delong, A., Weirauch, M.T. and Frey, B.J., 2015. Predicting the sequence specificities of DNA- and RNA-binding proteins by deep learning. Nature Biotechnology 33, 831-838.

AlQuraishi, M., 2019. End-to-end differentiable learning of protein structure. Cell Systems 8, 292-301.

Avsec, Ž., Weilert, M., Shrikumar, A., Alexandari, A., Krueger, S., Dalal, K., Fropf, R., McAnany, C., Gagneur, J., Kundaje, A. and Zeitlinger, J., 2019. Deep learning at base-resolution reveals motif syntax of the cis-regulatory code (bioRxiv).

Angermueller, C., Belanger, D., Gane, A., Mariet, Z., Dohan, D., Murphy, K., Colwell, L. and Sculley, D., 2020. Population-Based Black-Box Optimization for Biological Sequence Design (arXiv).

Arefeen, A., Xiao, X. and Jiang, T., 2019. DeepPASTA: deep neural network based polyadenylation site analysis. Bioinformatics, 35, 4577-4585.

Belanger, D., Vora, S., Mariet, Z., Deshpande, R., Dohan, D., Angermueller, C., Murphy, K., Chapelle, O. and Colwell, L., 2019. Biological Sequences Design using Batched Bayesian Optimization.

Bengio, Y., Léonard, N. and Courville, A., 2013. Estimating or propagating gradients through stochastic neurons for conditional computation (arXiv).

Bogard, N., Linder, J., Rosenberg, A. B. and Seelig, G., 2019. A Deep Neural Network for Predicting and Engineering Alternative Polyadenylation. Cell 178, 91-106.

Brookes, D.H., Park, H. and Listgarten, J., 2019. Conditioning by adaptive sampling for robust design (arXiv).

Butt, A.H., Alkhalaf, S., Iqbal, S. and Khan, Y.D., 2020. EnhancerP-2L: A Gene regulatory site identification tool for DNA enhancer region using CREs motifs (bioRxiv).

Chollet, F., 2015. Keras.

Cheng, J., Nguyen, T.Y.D., Cygan, K.J., Çelik, M.H., Fairbrother, W.G. \& Gagneur, J., 2019. MMSplice: modular modeling improves the predictions of genetic variant effects on splicing. Genome Biology, 20, 48.

Chung, J., Ahn, S. and Bengio, Y., 2016. Hierarchical multiscale recurrent neural networks (arXiv).

Costello, Z. and Martin, H.G., 2019. How to hallucinate functional proteins (arXiv).

Courbariaux, M., Hubara, I., Soudry, D., El-Yaniv, R. and Bengio, Y., 2016. Binarized neural networks: Training deep neural networks with weights and activations constrained to+ 1 or-1 (arXiv).

Deaton, R. J., Murphy, R. C., Garzon, M. H., Franceschetti, D. R. and Stevens Jr, S. E., 1996, June. Good encodings for DNA-based solutions to combinatorial problems. In DNA Based Computers, 247-258.

Di Giammartino, D.C., Nishida, K. and Manley, J.L., 2011. Mechanisms and consequences of alternative polyadenylation. Molecular cell, 43, 853-866.

Elkon, R., Ugalde, A.P. and Agami, R., 2013. Alternative cleavage and polyadenylation: extent, regulation and function. Nature Reviews Genetics, 14, 496-506.

Evans, R., Jumper, J., Kirkpatrick, J., Sifre, L., Green, T., Qin, C., Zidek, A., Nelson, A., Bridgland, A., Penedones, H. and Petersen, S., 2018. De novo structure prediction with deeplearning based scoring. Annu Rev Biochem, 77, 363-382.

Eraslan, G., Avsec, Ž., Gagneur, J. and Theis, F.J., 2019. Deep learning: new computational modelling techniques for genomics. Nature Reviews Genetics 20, 389-403.

Ernst, J., Melnikov, A., Zhang, X., Wang, L., Rogov, P., Mikkelsen, T.S. and Kellis, M., 2016. Genome-scale high-resolution mapping of activating and repressive nucleotides in regulatory regions. Nature biotechnology, 34, 1180-1190.

Gómez-Bombarelli, R., Wei, J.N., Duvenaud, D., Hernández-Lobato, J.M., Sánchez-Lengeling, B., Sheberla, D., Aguilera-Iparraguirre, J., Hirzel, T.D., Adams, R.P. and Aspuru-Guzik, A., 2018. Automatic chemical design using a data-driven continuous representation of molecules. ACS central science, 4, 268-276.

Gupta, A. and Zou, J., 2019. Feedback GAN for DNA optimizes protein functions. Nature Machine Intelligence 1, 105-111.

Hao, G. F., Xu, W. F., Yang, S. G. and Yang, G. F., 2015. Multiple simulated annealing-molecular dynamics (msa-md) for conformational space search of peptide and miniprotein. Scientific reports 5, 15568.

He, K., Zhang, X., Ren, S. and Sun, J., 2016. Deep residual learning for image recognition. In Proceedings of the IEEE conference on computer vision and pattern recognition, 770-778.

Ibrahim, Z., Khalid, N. K., Lim, K. S., Buyamin, S. and Mukred, J. A. A., 2011, December. A binary vector evaluated particle swarm optimization based method for DNA sequence design problem. In 2011 IEEE Student Conference on Research and Development, 160-164.

Jaganathan, K., Kyriazopoulou Panagiotopoulou, S., McRae, J. F., Darbandi, S. F., Knowles, D., Li, Y. I., … Farh, K. K.-H., 2019. Predicting Splicing from Primary Sequence with Deep Learning. Cell 176, 535-548.

Jang, E., Gu, S. and Poole, B., 2016. Categorical reparameterization with gumbel-softmax (arXiv).

Karollus, A., Avsec, Z. and Gagneur, J., 2020. Predicting Mean Ribosome Load for 5'UTR of any length using Deep Learning (bioRxiv).

Kheradpour, P. and Kellis, M., 2014. Systematic discovery and characterization of regulatory motifs in ENCODE TF binding experiments. Nucleic acids research, 42, 2976-2987.

Killoran, N., Lee, L. J., Delong, A., Duvenaud, D. and Frey, B. J., 2017. Generating and designing DNA with deep generative models (arXiv).

Kingma, D.P. and Welling, M., 2013. Auto-encoding variational bayes (arXiv).

Kingma, D. P. and Ba, J., 2014. Adam: A method for stochastic optimization (arXiv).

Kirkpatrick, S., Gelatt, C. D. and Vecchi, M. P., 1983. Optimization by simulated annealing. Science 220, 671-680.

Lakshminarayanan, B., Pritzel, A. and Blundell, C., 2017. Simple and scalable predictive uncertainty estimation using deep ensembles. Advances in neural information processing systems, 30, 6402-6413.

Lanchantin, J., Singh, R., Lin, Z. and Qi, Y., 2016. Deep motif: Visualizing genomic sequence classifications (arXiv).

Li, Z., Li, Y., Zhang, B., Li, Y., Long, Y., Zhou, J., Zou, X., Zhang, M., Hu, Y., Chen, W. and Gao, X., 2020. DeeReCT-APA: Prediction of Alternative Polyadenylation Site Usage Through Deep Learning (bioRxiv).

Linder, J., Bogard, N., Rosenberg, A.B. and Seelig, G., 2020. A generative neural network for maximizing fitness and diversity of synthetic DNA and protein sequences. Cell Systems, 11, 49-62.

Metropolis, N., Rosenbluth, A. W., Rosenbluth, M. N., Teller, A. H. and Teller, E., 1953. Equation of state calculations by fast computing machines. The Journal of Chemical Physics 21, 1087-1092.

Movva, R., Greenside, P., Marinov, G. K., Nair, S., Shrikumar, A. and Kundaje, A., 2019. Deciphering regulatory DNA sequences and noncoding genetic variants using neural network models of massively parallel reporter assays. PloS One, 14.

Mustaza, S. M., Abidin, A. F. Z., Ibrahim, Z., Shamsudin, M. A., Husain, A. R. and Mukred, J. A. A., 2011, December. A modified computational model of ant colony system in DNA sequence design. In 2011 IEEE Student Conference on Research and Development, 169-173.

Nguyen, Q.H., Nguyen-Vo, T.H., Le, N.Q.K., Do, T.T., Rahardja, S. and Nguyen, B.P., 2019. iEnhancer-ECNN: identifying enhancers and their strength using ensembles of convolutional neural networks. BMC genomics, 20, 951.

Norn, C., Wicky, B.I., Juergens, D., Liu, S., Kim, D., Koepnick, B., Anishchenko, I., Baker, D. and Ovchinnikov, S., 2020. Protein sequence design by explicit energy landscape optimization (bioRxiv).

Peters, J. and Schaal, S., 2007, June. Reinforcement learning by reward-weighted regression for operational space control. In Proceedings of the 24th international conference on Machine learning, 745-750.

Riesselman, A.J., Shin, J.E., Kollasch, A.W., McMahon, C., Simon, E., Sander, C., Manglik, A., Kruse, A.C. and Marks, D.S., 2019. Accelerating Protein Design Using Autoregressive Generative Models (bioRxiv).

Sample, P. J., Wang, B., Reid, D. W., Presnyak, V., McFadyen, I. J., Morris, D. R. and Seelig, G., 2019. Human 5' UTR design and variant effect prediction from a massively parallel translation assay. Nature Biotechnology 37, 803-809.

Sandelin, A., Alkema, W., Engström, P., Wasserman, W.W. and Lenhard, B., 2004. JASPAR: an open‐access database for eukaryotic transcription factor binding profiles. Nucleic acids research, 32, D91-D94.

Schreiber, J., Lu, Y.Y. and Noble, W.S., 2020. Ledidi: Designing genome edits that induce functional activity (bioRxiv).

Senior, A.W., Evans, R., Jumper, J., Kirkpatrick, J., Sifre, L., Green, T., Qin, C., Žídek, A., Nelson, A.W., Bridgland, A. and Penedones, H., 2020. Improved protein structure prediction using potentials from deep learning. Nature, 577, 706-710.

Shi, Y., 2012. Alternative polyadenylation: new insights from global analyses. Rna, 18, 2105-2117.

Simonyan, K., Vedaldi, A. and Zisserman, A., 2013. Deep inside convolutional networks: Visualising image classification models and saliency maps (arXiv).

Tareen, A. and Kinney, J.B., 2019. Biophysical models of cis-regulation as interpretable neural networks (arXiv).

Tian, B. and Manley, J.L., 2017. Alternative polyadenylation of mRNA precursors. Nature reviews Molecular cell biology, 18, 18-30.

Tischer, D., Lisanza, S., Wang, J., Dong, R., Anishchenko, I., Milles, L.F., Ovchinnikov, S. and Baker, D., 2020. Design of proteins presenting discontinuous functional sites using deep learning (bioRxiv).

Ulyanov, D., Vedaldi, A. and Lempitsky, V., 2016. Instance normalization: The missing ingredient for fast stylization (arXiv).

Xiao, J., Xu, J., Chen, Z., Zhang, K. and Pan, L., 2009. A hybrid quantum chaotic swarm evolutionary algorithm for DNA encoding. Computers \& Mathematics with Applications 57, 1949-1958.

Yang, K.K., Wu, Z. and Arnold, F.H., 2019. Machine-learning-guided directed evolution for protein engineering. Nature Methods 16, 687-694.

Yang, J., Anishchenko, I., Park, H., Peng, Z., Ovchinnikov, S. and Baker, D., 2020. Improved protein structure prediction using predicted interresidue orientations. Proceedings of the National Academy of Sciences.

Zhou, J. and Troyanskaya, O. G., 2015. Predicting effects of noncoding variants with deep learning–based sequence model. Nature Methods 12, 931-934.

Zou, J., Huss, M., Abid, A., Mohammadi, P., Torkamani, A. and Telenti, A., 2019. A primer on deep learning in genomics. Nature genetics 51, 12-18.

\newpage

\normalsize

\section*{Supplementary Information}

\setcounter{figure}{0}

\renewcommand{\thefigure}{S\arabic{figure}}

\section*{Appendix A: Supplemental Activation Maximization Comparisons}

\subsection*{Comparison to Evolutionary Algorithms and Simulated Annealing}

We compared Fast SeqProp to the discrete nucleotide-swapping search algorithm 'Evolution' (Sample et al., 2019; see Methods) as well as the global optimization heuristic Simulated Annealing (Figure S1A; Kirkpatrick et al., 1983). We benchmarked the methods on the DragoNN maximization task. The results show that Fast SeqProp significantly outperforms Simulated Annealing (Figure S1B); the best fitness score that Simulated Annealing reached in $20,000$ iterations was reached by Fast SeqProp in less than $1,000$ iterations, marking a 20x speed-up. Interestingly, increasing the number of substitutions at each step of Simulated Annealing initially increases the rate of convergence, but results in worse optima. Furthermore, the pairwise nucleotide-swapping search, Evolution, never makes any improvement on its randomly initialized start sequence, suggesting that pair-wise nucleotide changes is not powerful enough to maximize DragoNN.

\subsection*{Pathologies of Softmax Relaxation}

We hypothesized that some predictors, having been trained only on discrete one-hot coded patterns, may have poor predictive power on continuous-valued softmax sequence relaxations (input to PWM). To test this, we measured both the 'training' loss $\mathcal{L}_{\text{train}}$ (which is based on the softmax sequence input $\sigma(\bm{l})$ for PWM and discrete samples $\delta(\bm{l})$ for Fast SeqProp) and test loss $\mathcal{L}_{\text{test}}$ (which is based on discrete samples $\delta(\bm{l})$ regardless of design method) when maximizing MPRA-DragoNN.

Indeed, maximizing MPRA-DragoNN with PWM leads to an overestimated predictor score on the softmax input (Figure S1C; top), as the training loss is more than 6-fold lower than the test loss. Using Fast SeqProp, on the other hand, the training and test losses are identical (Figure S1C; bottom). While the training loss is 2x higher than the training loss of PWM, the test loss is more than 3x lower.

\subsection*{Entropy Penalties and the Gumbel Distribution}

Curious whether the gap observed between training and test loss for the PWM method in Figure S1C was caused by high softmax entropy, we tested whether an explicit entropy penalty, $\lambda \cdot \frac{1}{N} \sum_{i=1}^{N} \sum_{j=1}^{M} - \sigma(\bm{l})_{ij} \cdot \log_{2} \sigma(\bm{l})_{ij}$, would improve the method. We re-optimized sequences for Optimus 5' and DragoNN, such that the mean nucleotide conservation reached at least $1.5 / 2.0$ bits (Figure S1D). Even at low entropy, PWM does not converge to as good minima as Fast SeqProp.

We also compared the performance of our logit-normalized, softmax straight-through design method Fast SeqProp to a version of the method using the Gumbel distribution for sampling (Jang et al., 2016; temperature $\tau = 0.1$; Figure S1E). While the Gumbel variant of the design method reached the same optima as Fast SeqProp, it converged slower. Importantly, same as PWM and SeqProp, the Gumbel design method benefited substantially from logit normalization.

\subsection*{Insensitivity to Optimizer Settingss}

We noted that the performance of the PWM method was dependent on optimizer settings (Figure S1F); the speed at which it could maximize APARENT was increased by switching to an SGD optimizer and setting a very high learning rate. However, the method could still not reach the same optimum as Fast SeqProp, which operated well at a default SGD learning rate of $0.1$.

\subsection*{Superiority of Softmax Straight-Through Gradients}

We compared the performance of Fast SeqProp, which uses the softmax straight-through gradient estimator $\frac{\partial \delta(\bm{l})_{ij}}{\partial \bm{l}_{ik}} = \frac{\partial \sigma(\bm{l})_{ij}}{\partial \bm{l}_{ik}} = \sigma(\bm{l})_{ik} \cdot (\mathbbm{1}_{(j = k)} - \sigma(\bm{l})_{ij})$, to a version using the original estimator $\frac{\partial \delta(\bm{l})_{ij}}{\partial \bm{l}_{ij}} = 1$. As demonstrated on DragoNN, the softmax estimator reaches much better optima (Figure S1G). Sampling multiple sequences $\{\delta(\bm{l})^{(s)}\}_{s=1}^{S}$ at each update and walking down the average gradient $\frac{1}{S} \sum_{s=1}^{S} \nabla_{\bm{l}} \mathcal{P}(\delta(\bm{l})^{(s)})$ slightly speeds up convergence, but does not improve optima.

\begin{figure}[H]
  \centering
  \includegraphics[scale=0.85]{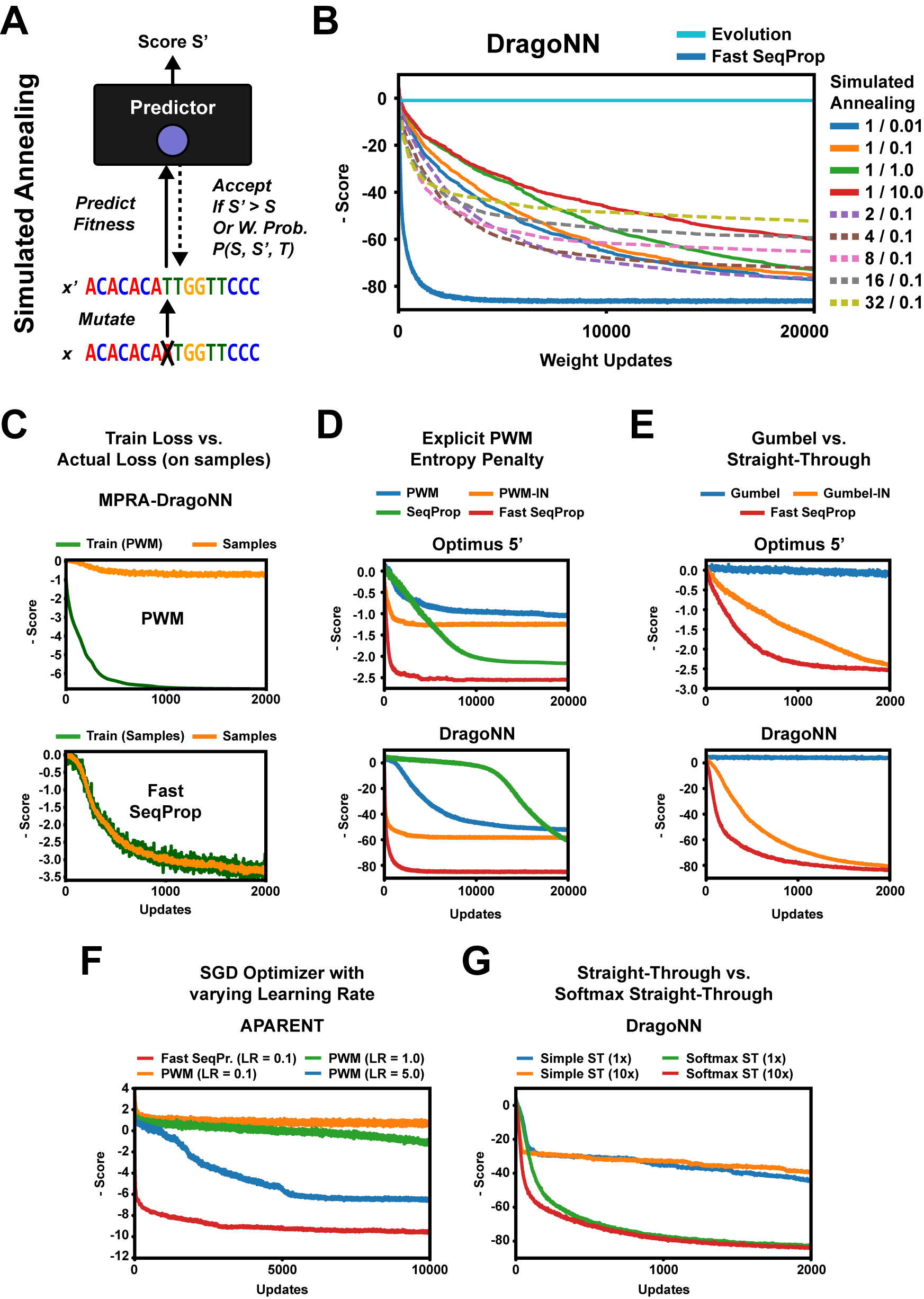}
  \caption{ (A) In Simulated Annealing, mutations are accepted with a temperature-controlled probability even if the predicted fitness decreases. (B) Maximizing DragoNN SPI1. Simulated Annealing is tested at several parameter configurations (number of substitutions per step / initial temperature). (C) Maximizing MPRA-DragoNN using either (top) PWM or (bottom) Fast SeqProp as the design method. Shown are the optimization scores during the design phase (green) when we pass either the softmax sequence (PWM) or a sampled sequence (Fast SeqProp) to the predictor, and the corresponding scores if we were to sample sequences from the softmax representation (orange). (D) Maximizing Optimus 5' and DragoNN with a softmax sequence entropy penalty. (E) Comparing Fast SeqProp to a version of the same method using the Gumbel distribution ('Gumbel' -- Gumbel-sampling, 'Gumbel-IN' -- Gumbel-sampling with instance norm.). (F) Maximizing APARENT using Fast SeqProp or PWM with SGD (LR = Learning Rate). (G) Maximizing DragoNN using Fast SeqProp, with either the softmax- or original (simple) straight-through estimator. 1x and 10x refer to the number of sequences sampled at each update. }
  \label{fig:s1}
\end{figure}

\section*{Appendix B: Additional Sequence Optimization Examples}

Figure S2 shows an example optimization run for each predictor, comparing the softmax sequences $\sigma(\bm{l})$ generated by the PWM and Fast SeqProp design methods. The same sequences as those shown in Figure \ref{fig:2} are also shown here, but for $200$, $2,000$ and $20,000$ iterations of gradient ascent. The logit matrices $\bm{l}$ were uniformly randomly initialized and default Adam parameters were used.

\begin{figure}[H]
  \centering
  \includegraphics[scale=0.95]{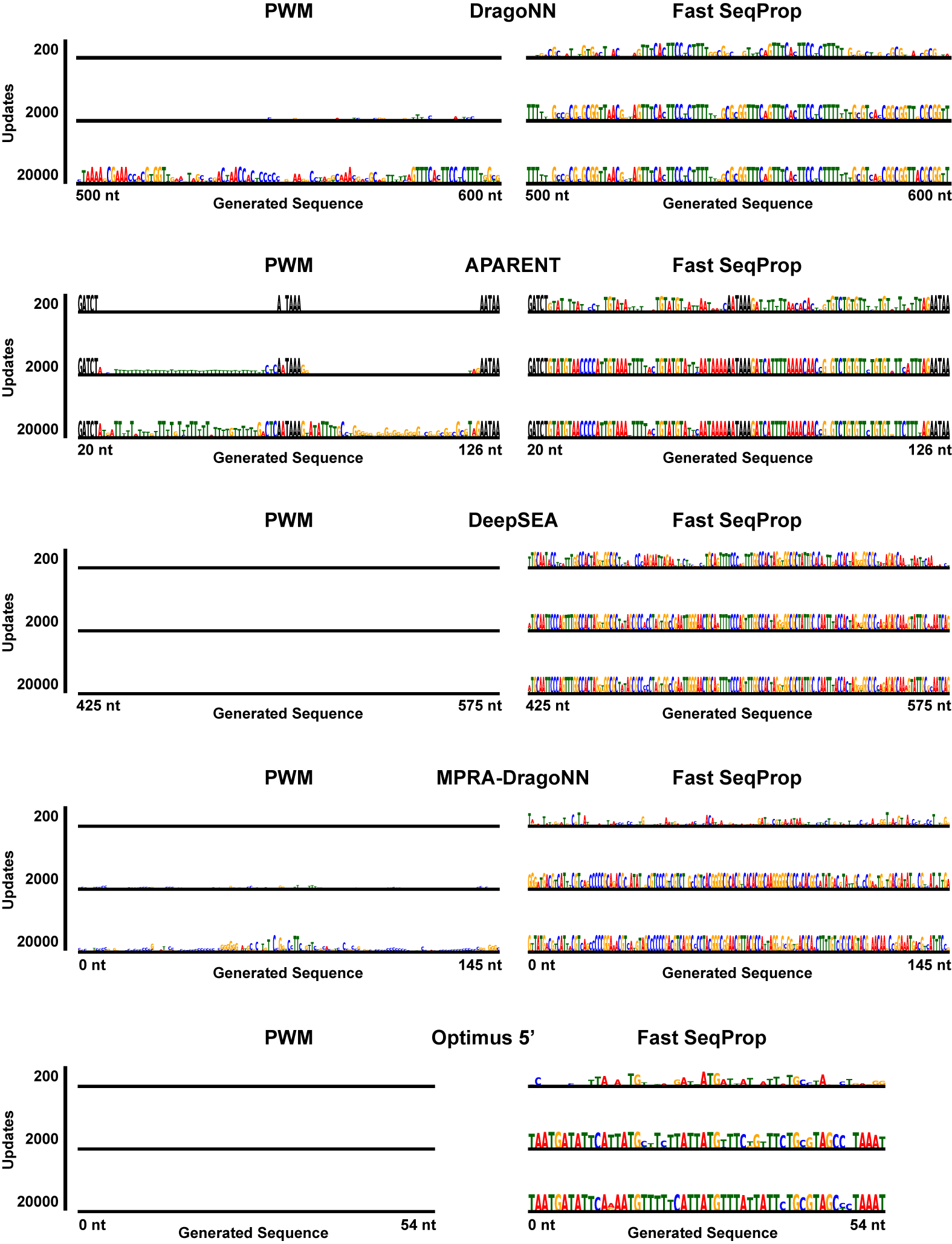}
  \caption{ Generation of sequences which maximize the score of DragoNN (SPI1), APARENT, DeepSEA (CTCF, Dnd41), MPRA-DragoNN (SV40 Mean activity) and Optimus 5'. Shown are the resulting sequence logos of running the PWM and Fast SeqProp optimization methods for $200$, $2,000$ and $20,000$ iterations respectively, starting from randomly initialized logits. }
  \label{fig:s2}
\end{figure}

\section*{Appendix C: Additional Sequence Design Comparisons}

In Figure S3A-D we report additional performance metrics tracked for each of the benchmarked methods when designing sequences for APARENT (polyadenylation signals), MPRA-DragoNN (transcriptional enhancers) and Optimus 5' (translationally efficient 5' UTRs). We measure median edit distance between designed sequences in Figure S3B as a proxy for diversity. In Figure S3C we measure the strength (relative probability of selection) of designed polyadenylation signals using the held-out model DeepPASTA. In Figure S3D, we measure the strength of designed enhancer sequences using the held-out model EnhancerP-2L.

In Figure S3E we perform a benchmark analysis comparing Fast SeqProp to the methods CbAS (Brookes et al., 2019), AM-VAE (Killoran et al., 2017), FB-VAE (Gupta et al., 2019) and RWR (Peters et al., 2007). The benchmark is very similar to the one presented in the main paper (Figure \ref{fig:3}B), but instead of using the pre-trained oracle predictors APARENT, MPRA-DragoNN and Optimus 5' we here train new probabilistic oracles based on the loss function from Lakshminarayanan et al. (2017). These models are capable of estimating the uncertainty in their fitness predictions, a property that can be used to regularize the sequence optimization based on predictor confidence. Specifically, for Fast SeqProp, we use the (differentiable) survival function of the normal distribution in order to maximize the probability that the predicted fitness of designed sequences is larger than a particular quantile $q$ of the training data (see Equation \ref{eq:balaji-vae-seqprop} in Methods). The survival function is also used as part of the sample re-weighting scheme in CbAS (Brookes et al., 2019). Here, we use $q = 0.95$.

We carried out the benchmark on three design tasks: (1) Polyadenylation (pA) signal design (by training the oracle on the same data as APARENT was trained on), (2) 5' UTR design (by training the oracle on the same data as Optimus 5') and (3) Green Fluorescent Protein (GFP) variant design (replicating the test suite from Brookes et al., 2019). For pA signal design and 5' UTR design, we used a single oracle model for each design task, consisting of two convolutional layers and a single fully connected hidden layer. After training, the oracle predictions were highly correlated with measured fitness on held-out data (pearson $r=0.87$ and $0.91$ for each respective task). For the GFP design task, we used an ensemble of 5 predictors (each consisting of a single fully connected hidden layer). We used DeeReCT-APA to validate the pA design task (Li et al., 2020). We used the retrained version of Optimus 5' to validate the 5' UTRs (Sample et al., 2019). Finally, we used the same GP regression model as was used by Brookes et al. (2019) to validate the GFP task. When running CbAS and FB-VAE, we used $q=0.8$ as the quantile cutoff. For CbAS, FB-VAE and RWR, we used 1000 samples per feedback round except for the GFP task where 100 samples were used. The same VAE models that were used in the main paper were also used here for pA signal and 5' UTR design.

As shown in Figure S3E, for the pA signal task, Fast SeqProp designs sequences with higher validation scores than all other methods. Both Fast SeqProp and AM-VAE converge to higher scores than CbAS, FB-VAE and RWR using $100$x-$1,000$x fewer calls to the oracle. For the 5' UTR design task, RWR ultimately reaches marginally higher validation score than Fast SeqProp, but does so after $100$x more oracle calls. Fast SeqProp reaches higher validation scores than all other methods. Interestingly, for AM-VAE (which is also based on activation maximization), the validation scores decrease throughout the course of optimization. Finally, for the GFP task, methods AM-VAE, CbAS, FB-VAE and RWR initially start with significantly higher oracle- and validation scores than Fast SeqProp. The reason is that these methods start by sampling sequences from the pre-trained VAE, which already has high validation scores. Throughout the course of optimization, these methods marginally increase their validation scores. However, after $\sim 1,000$ predictor calls, Fast SeqProp quickly converges to sequence designs with higher validation scores than all other methods.

We tried using the above approach on the MPRA-DragoNN enhancer data from Ernst et al., (2016), but we failed to train a sufficiently accurate oracle predictor with the loss function from Lakshminarayanan et al. (2017). This is likely due to the substantial levels of noise present in the training data (as is also described by Movva et al., 2019). This enhancer dataset is unique in that it contains replicate functional measurements in multiple cell types (K562 and HepG2) and in theory would allow the design of cell-type specific enhancer sequences. However, if we cannot even fit the oracle model to individual cell type measurements, we have no hope of fitting the oracle to differences between two (noisy) cell type readouts (which would enable cell-type specific design). When we attempted this, we ended up with a very weak oracle (pearson $r=0.05$ on held-out data, using the difference in measured transcriptional activity between HepG2 and K562 with SV40 promoters as target value).

\begin{figure}[H]
  \centering
  \includegraphics[scale=0.78]{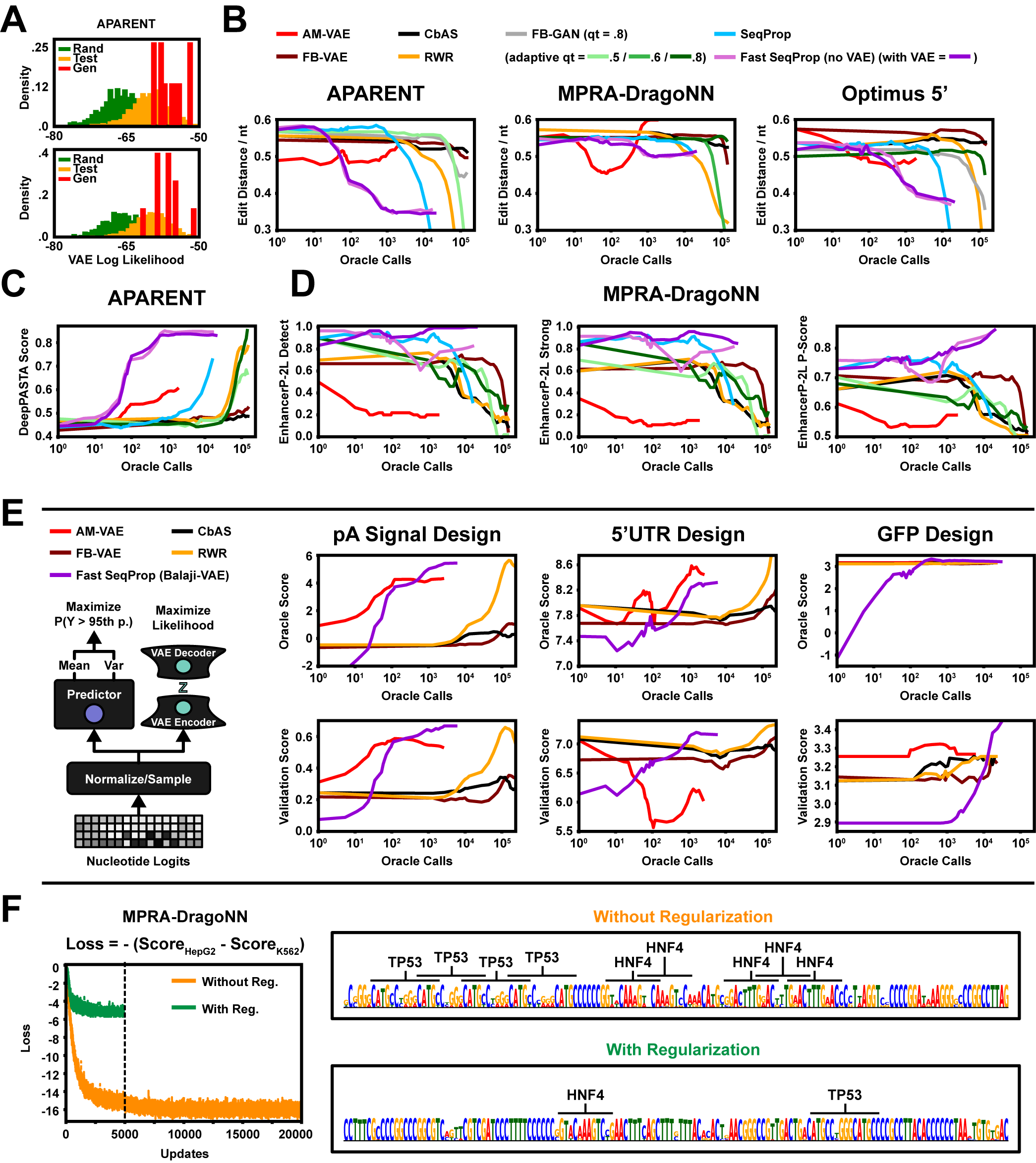}
  \caption{ (A) Estimated VAE log likelihood distribution of random sequences (green), test sequences from the APARENT dataset (orange) and designed sequences (red), using Fast SeqProp without (top) and with VAE regularization (bottom). (B) Sequence diversity of each design methods as a function of predictor calls, measured as the median relative edit distance between sequence samples per method, across three repeats. (C) Median predicted relative use of the designed polyadenylation signals for each method, using the DeepPASTA model. Median score reported across three repeats. (D) Three predicted types of validation scores using the EnhancerP-2L model: (left) The mean frequency of identified functional enhancers, (middle) the mean weighted frequency of identified enhancers (weak enhancers are worth 0.5, strong enhancer 1.0), and (right) Predicted EnhancerP-2L P-scores. The means are computed across three repeats. (E) Additional benchmark analysis using both VAE-regularization and a probabilistic oracle ensemble capable of estimating uncertainty (Lakshminarayanan et al., 2017; see Appendix C for details). Top: Median oracle fitness scores across three repeats for the tasks: pA signal design, 5' UTR design and GFP variant design. Bottom: Median validation scores across three repeats, using DeeReCT-APA, Optimus 5' (retrained) and a GP regression model respectively. (F) Maximizing the difference in transcriptional activity between K562 and HepG2. Shown are the loss and an example designed sequence when running Fast SeqProp without regularization (orange) and when applying both VAE- and Activity-regularization (green). }
  \label{fig:s3}
\end{figure}

Instead, we used the original MPRA-DragoNN predictor, but with regularization penalties applied to the sum of activations across some of the model's internal convolutional layers in order to design robust, HepG2-specific enhancers (see Equation \ref{eq:act-reg-vae-seqprop} in Methods). Specifically, we maximized the predicted cell-type specificity $\mathcal{P}(\bm{x})^{(\text{HepG2})} - \mathcal{P}(\bm{x})^{(\text{K562})}$ while also minimizing a margin loss applied to the sum of the $5$:th, $8$:th and $9$:th convolutional activation maps (and also minimizing the VAE-loss as before). Figure S3F shows the result of using Fast SeqProp to design maximally HepG2-specific enhancers, with and without regularization. Without regularization, the design method fills an example designed sequence with $4$ TP53 binding motifs and $5$ HNF4 binding motifs. With regularization, only one copy of each motif appears in the designed example.

\section*{Appendix D: Extra Protein Structure Optimization Example}

\begin{figure}[H]
  \centering
  \includegraphics[scale=0.85]{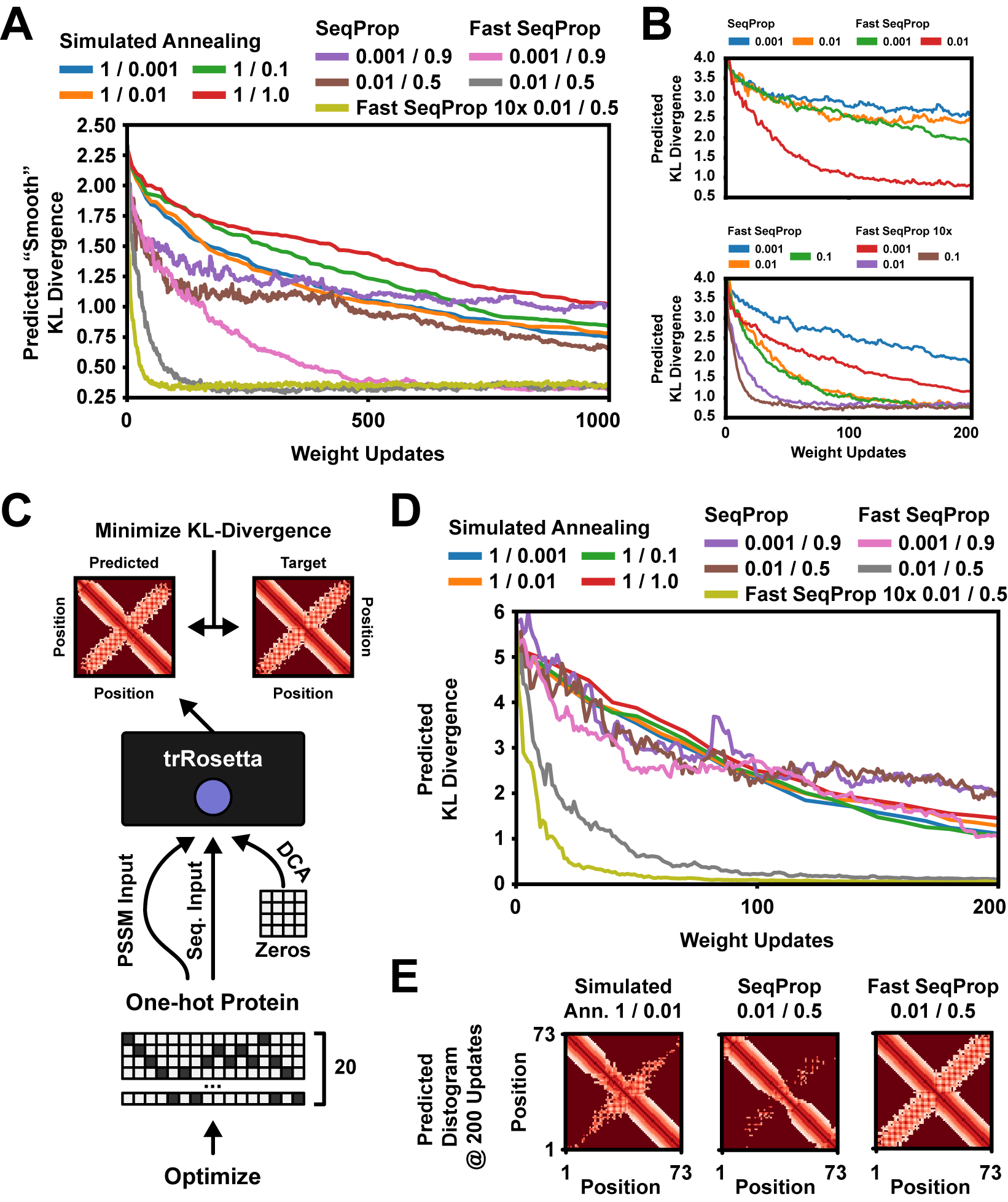}
  \caption{ (A) Identical optimization as the one shown in Figure \ref{fig:4}B, but here we report the "Smooth" KL divergence, as described in Appendix D. (B) The effect of  various Adam learning rate- and momentum parameter settings during structure optimization. "10x" refers to sampling 10 sequences at each step of Fast SeqProp and walking down the average gradient. (C) Sequences are designed to conform to a coiled-coil hairpin structure, as predicted by trRosetta. (D) Sequence optimization results after 1000 iterations. Simulated Annealing was tested at several initial temperatures and SeqProp / Fast SeqProp was tested with multiple optimizer parameters (Adam). (E) Predicted residue distance distributions after 200 iterations. }
  \label{fig:s4}
\end{figure}

The categorical KL-divergence measured in Figure \ref{fig:4}B of the main paper is not necessarily a good metric for estimating the distance between an optimized protein structure and its target structure. For example, if the optimized structure puts all of its probability mass just one discretized bin away in the predicted distance map $\bm{D}^{P}_{ijk+1}$ compared to the target distance map $\bm{D}^{T}_{ijk}$, the KL-divergence becomes nearly maximal even though the actual distances between structures is quite small.

To mitigate this issue, we validate the optimized structures using a 'smooth' KL-divergence metric, which transforms the discretized, binned distance maps $\bm{D}$ and angle distributions $\bm{\theta}, \bm{\omega}, \bm{\phi}$ into a single weighted probability at each position $i, j$. For $\bm{\theta}$ and $\bm{\omega}$, we use sin- and cos transforms in order to take into account that bins $\bm{\theta}_{ij1}$ and $\bm{\theta}_{ijK}$ are close in the unit circle. We also take into account that, in the distance and angle distributions predicted by trRosetta, bin 0 has the special meaning 'no contact'. The smooth KL-divergence $\mathcal{L}_{\text{smooth}}\big(\bm{D}^{P}, \bm{\theta}^{P}, \bm{\omega}^{P}, \bm{\phi}^{P}\big)$ is computed according to the following formulas:

\begin{align*}
    \mathcal{L}_{\text{smooth}}\big(\bm{D}^{P}, \bm{\theta}^{P}, \bm{\omega}^{P}, \bm{\phi}^{P}\big) &= \text{KL}_{\text{smooth}}(\bm{D}^{P} || \bm{D}^{T}) + \text{KL}_{\text{smooth}}(\bm{\phi}^{P} || \bm{\phi}^{T})\\
    &+ \text{KL}_{\text{circular}}(\bm{\theta}^{P} || \bm{\theta}^{T}) + \text{KL}_{\text{circular}}(\bm{\omega}^{P} || \bm{\omega}^{T})\\
    \text{KL}_{\text{smooth}}(\bm{X} || \bm{Y}) &= \frac{1}{N^{2}} \cdot \sum_{i=1}^{N} \sum_{j=1}^{N} \bigg(\text{Smooth}(\bm{Y})_{ij} \cdot \log \bigg( \frac{\text{Smooth}(\bm{Y})_{ij}}{\text{Smooth}(\bm{X})_{ij}} \bigg)\\
    &\quad \quad \quad \quad + \bm{Y}_{ij0} \cdot \log \bigg( \frac{\bm{Y}_{ij0}}{\bm{X}_{ij0}} \bigg) + \bm{\epsilon}(\bm{Y})_{ij} \cdot \log \bigg( \frac{\bm{\epsilon}(\bm{Y})_{ij}}{\bm{\epsilon}(\bm{X})_{ij}} \bigg) \bigg)\\
    \text{KL}_{\text{circular}}(\bm{X} || \bm{Y}) &= \frac{1}{N^{2}} \cdot \sum_{i=1}^{N} \sum_{j=1}^{N} \bigg(\text{Circular}(\bm{Y})_{ij} \cdot \log \bigg( \frac{\text{Circular}(\bm{Y})_{ij}}{\text{Circular}(\bm{X})_{ij}} \bigg)\\
    &\quad \quad \quad \quad + \bm{Y}_{ij0} \cdot \log \bigg( \frac{\bm{Y}_{ij0}}{\bm{X}_{ij0}} \bigg) + \bm{\eta}(\bm{Y})_{ij} \cdot \log \bigg( \frac{\bm{\eta}(\bm{Y})_{ij}}{\bm{\eta}(\bm{X})_{ij}} \bigg) \bigg)
\end{align*}
\begin{align*}
    \text{Smooth}(\bm{X})_{ij} &= \sum_{k=1}^{K} \bigg(\frac{k-1}{K-1}\bigg) \cdot \bm{X}_{ijk}\\
    \text{Circular}(\bm{X})_{ij} &= \sum_{k=1}^{K} 0.5 \cdot \bigg[\bigg(\text{sin}\bigg[\bigg(\frac{k-1}{K-1}\bigg) \cdot 2\pi - \pi \bigg] \cdot 0.5 + 0.5\bigg) \cdot \bm{X}_{ijk}\bigg]\\
    &+ \sum_{k=1}^{K} 0.5 \cdot \bigg[\bigg(\text{cos}\bigg[\bigg(\frac{k-1}{K-1}\bigg) \cdot 2\pi - \pi \bigg] \cdot 0.5 + 0.5\bigg) \cdot \bm{X}_{ijk}\bigg]\\
    \bm{\epsilon}(\bm{X})_{ij} &= 1 - \text{Smooth}(\bm{X})_{ij} - \bm{X}_{ij0}\\
    \bm{\eta}(\bm{X})_{ij} &= 1 - \text{Circular}(\bm{X})_{ij} - \bm{X}_{ij0}
\end{align*}

In addition to the protein structure design task evaluated in Figure \ref{fig:4}, we also benchmarked SeqProp, Fast SeqProp and Simulated Annealing on a separate protein structure. Here, we task the methods with designing sequences conforming to a coiled-coil hairpin structure, again using trRosetta as the differentiable structure predictor (Yang et al., 2020). The same KL-divergence loss as was used in Figure \ref{fig:4} was used here. The results, depicted in Figure S4C-E, show that Fast SeqProp converges very quickly to a near-optimal (zero) KL-divergence. The structure is likely easier to design sequences for, as there is only one major long-ranging contact formation. The sequence is also only about half as long as the one in Figure \ref{fig:4}.

\end{document}